\documentstyle[12pt,fullpage,chicagob,defn,url]{article}
\begin{document}

 
\newtheorem{THEOREM}{Theorem}[section]
\newenvironment{theorem}{\begin{THEOREM} \hspace{-.85em} {\bf :} }%
                        {\end{THEOREM}}
\newtheorem{LEMMA}[THEOREM]{Lemma}
\newenvironment{lemma}{\begin{LEMMA} \hspace{-.85em} {\bf :} }%
                      {\end{LEMMA}}
\newtheorem{COROLLARY}[THEOREM]{Corollary}
\newenvironment{corollary}{\begin{COROLLARY} \hspace{-.85em} {\bf :} }%
                          {\end{COROLLARY}}
\newtheorem{PROPOSITION}[THEOREM]{Proposition}
\newenvironment{proposition}{\begin{PROPOSITION} \hspace{-.85em} {\bf :} }%
                            {\end{PROPOSITION}}
\newtheorem{DEFINITION}[THEOREM]{Definition}
\newenvironment{definition}{\begin{DEFINITION} \hspace{-.85em} {\bf :} \rm}%
                            {\end{DEFINITION}}
\newtheorem{CLAIM}[THEOREM]{Claim}
\newenvironment{claim}{\begin{CLAIM} \hspace{-.85em} {\bf :} \rm}%
                            {\end{CLAIM}}
\newtheorem{EXAMPLE}[THEOREM]{Example}
\newenvironment{example}{\begin{EXAMPLE} \hspace{-.85em} {\bf :} \rm}%
                            {\end{EXAMPLE}}
\newtheorem{REMARK}[THEOREM]{Remark}
\newenvironment{remark}{\begin{REMARK} \hspace{-.85em} {\bf :} \rm}%
                            {\end{REMARK}}
 
\newcommand{\thm}{\begin{theorem}}
\newcommand{\lem}{\begin{lemma}}
\newcommand{\pro}{\begin{proposition}}
\newcommand{\dfn}{\begin{definition}}
\newcommand{\rem}{\begin{remark}}
\newcommand{\xam}{\begin{example}}
\newcommand{\cor}{\begin{corollary}}
\newcommand{\prf}{\noindent{\bf Proof:} }
\newcommand{\ethm}{\end{theorem}}
\newcommand{\elem}{\end{lemma}}
\newcommand{\epro}{\end{proposition}}
\newcommand{\edfn}{\bbox\end{definition}}
\newcommand{\erem}{\bbox\end{remark}}
\newcommand{\exam}{\bbox\end{example}}
\newcommand{\ecor}{\end{corollary}}
\newcommand{\eprf}{\bbox\vspace{0.1in}}
\newcommand{\beqn}{\begin{equation}}
\newcommand{\eeqn}{\end{equation}}
\newcommand{\wbox}{\mbox{$\sqcap$\llap{$\sqcup$}}}
\newcommand{\bbox}{\vrule height7pt width4pt depth1pt}
\newcommand{\qed}{\eprf}
\newcommand{\clm}{\begin{claim}}
\newcommand{\eclm}{\end{claim}}
\let\member=\in
\let\notmember=\notin
\newcommand{\sub}{_}
\def\su{^}
\newcommand{\rarrow}{\rightarrow}
\newcommand{\larrow}{\leftarrow}
\newcommand{\boldsymbol}[1]{\mbox{\boldmath $\bf #1$}}
\newcommand{\bolda}{{\bf a}}
\newcommand{\boldb}{{\bf b}}
\newcommand{\boldc}{{\bf c}}
\newcommand{\boldd}{{\bf d}}
\newcommand{\bolde}{{\bf e}}
\newcommand{\boldf}{{\bf f}}
\newcommand{\boldg}{{\bf g}}
\newcommand{\boldh}{{\bf h}}
\newcommand{\boldi}{{\bf i}}
\newcommand{\boldj}{{\bf j}}
\newcommand{\boldk}{{\bf k}}
\newcommand{\boldl}{{\bf l}}
\newcommand{\boldm}{{\bf m}}
\newcommand{\boldn}{{\bf n}}
\newcommand{\boldo}{{\bf o}}
\newcommand{\boldp}{{\bf p}}
\newcommand{\boldq}{{\bf q}}
\newcommand{\boldr}{{\bf r}}
\newcommand{\bolds}{{\bf s}}
\newcommand{\boldt}{{\bf t}}
\newcommand{\boldu}{{\bf u}}
\newcommand{\boldv}{{\bf v}}
\newcommand{\boldw}{{\bf w}}
\newcommand{\boldx}{{\bf x}}
\newcommand{\boldy}{{\bf y}}
\newcommand{\boldz}{{\bf z}}
\newcommand{\boldA}{{\bf A}}
\newcommand{\boldB}{{\bf B}}
\newcommand{\boldC}{{\bf C}}
\newcommand{\boldD}{{\bf D}}
\newcommand{\boldE}{{\bf E}}
\newcommand{\boldF}{{\bf F}}
\newcommand{\boldG}{{\bf G}}
\newcommand{\boldH}{{\bf H}}
\newcommand{\boldI}{{\bf I}}
\newcommand{\boldJ}{{\bf J}}
\newcommand{\boldK}{{\bf K}}
\newcommand{\boldL}{{\bf L}}
\newcommand{\boldM}{{\bf M}}
\newcommand{\boldN}{{\bf N}}
\newcommand{\boldO}{{\bf O}}
\newcommand{\boldP}{{\bf P}}
\newcommand{\boldQ}{{\bf Q}}
\newcommand{\boldR}{{\bf R}}
\newcommand{\boldS}{{\bf S}}
\newcommand{\boldT}{{\bf T}}
\newcommand{\boldU}{{\bf U}}
\newcommand{\boldV}{{\bf V}}
\newcommand{\boldW}{{\bf W}}
\newcommand{\boldX}{{\bf X}}
\newcommand{\boldY}{{\bf Y}}
\newcommand{\boldZ}{{\bf Z}}
\newcommand{\sat}{\models}
\newcommand{\dtur}{\models}
\newcommand{\infers}{\vdash}
\newcommand{\stur}{\vdash}
\newcommand{\rimp}{\Rightarrow}
\newcommand{\limp}{\Leftarrow}
\newcommand{\dimp}{\Leftrightarrow}
\newcommand{\bor}{\bigvee}
\newcommand{\band}{\bigwedge}
\newcommand{\union}{\cup}
\newcommand{\inter}{\cap}
\newcommand{\xx}{{\bf x}}
\newcommand{\yy}{{\bf y}}
\newcommand{\uu}{{\bf u}}
\newcommand{\vv}{{\bf v}}
\newcommand{\FF}{{\bf F}}
\newcommand{\natnum}{{\sl N}}
\newcommand{\IR}{\mbox{$I\!\!R$}}
\newcommand{\IP}{\mbox{$I\!\!P$}}
\newcommand{\IN}{\mbox{$I\!\!N$}}
\newcommand{\IC}{\mbox{$C\!\!\!\!\raisebox{.75pt}{\mbox{\sqi I}}$}}
\newcommand{\marrow}{\hbox{$\rightarrow$ \hskip -10pt
                      $\rightarrow$ \hskip 3pt}}
\renewcommand{\phi}{\varphi}
\newcommand{\Circ}{\mbox{{\small $\bigcirc$}}}
\newcommand{\lt}{<}
\newcommand{\gt}{>}
\newcommand{\all}{\forall}
\newcommand{\infinity}{\infty}
\newcommand{\bc}[2]{\left( \begin{array}{c} #1 \\ #2 \end{array} \right)}
\newcommand{\cross}{\times}
\newcommand{\bigfootnote}[1]{{\footnote{\normalsize #1}}}
\newcommand{\medfootnote}[1]{{\footnote{\small #1}}}
\newcommand{\bd}{\bf}
 
 
\newcommand{\imp}{\Rightarrow}
 
\newcommand{\A}{{\cal A}}
\newcommand{\B}{{\cal B}}
\newcommand{\C}{{\cal C}}
\newcommand{\D}{{\cal D}}
\newcommand{\E}{{\cal E}}
\newcommand{\F}{{\cal F}}
\newcommand{\G}{{\cal G}}
\newcommand{\I}{{\cal I}}
\newcommand{\J}{{\cal J}}
\newcommand{\K}{{\cal K}}
\newcommand{\M}{{\cal M}}
\newcommand{\N}{{\cal N}}
\newcommand{\Ocal}{{\cal O}}
\newcommand{\Hcal}{{\cal H}}
\renewcommand{\P}{{\cal P}}
\newcommand{\Q}{{\cal Q}}
\newcommand{\R}{{\cal R}}
\newcommand{\T}{{\cal T}}
\newcommand{\U}{{\cal U}}
\newcommand{\V}{{\cal V}}
\newcommand{\W}{{\cal W}}
\newcommand{\X}{{\cal X}}
\newcommand{\Y}{{\cal Y}}
\newcommand{\Z}{{\cal Z}}
 
\newcommand{\Kone}{{\cal K}_1}
\newcommand{\abs}[1]{\left| #1\right|}
\newcommand{\set}[1]{\left\{ #1 \right\}}
\newcommand{\Ki}{{\cal K}_i}
\newcommand{\Kn}{{\cal K}_n}
\newcommand{\st}{\, \vert \,} 
\newcommand{\stc}{\, : \,} 
\newcommand{\la}{\langle}
\newcommand{\ra}{\rangle}
\newcommand{\<}{\langle}
\renewcommand{\>}{\rangle}
\newcommand{\lang}{\mbox{${\cal L}_n$}}
\newcommand{\langd}{\mbox{${\cal L}_n^D$}}

\newtheorem{nlem}{Lemma}
\newtheorem{Ob}{Observation}
\newtheorem{pps}{Proposition}
\newtheorem{defn}{Definition}
\newtheorem{crl}{Corollary}
\newtheorem{cl}{Claim}
\newcommand{\pf}{\par\noindent{\bf Proof}~~}
\newcommand{\eg}{e.g.,~}
\newcommand{\ie}{i.e.,~}
\newcommand{\vs}{vs.~}
\newcommand{\cf}{cf.~}
\newcommand{\etal}{et al.\ }
\newcommand{\resp}{resp.\ }
\newcommand{\respc}{resp.,\ }
\newcommand{\comment}[1]{\marginpar{\scriptsize\raggedright #1}}
\newcommand{\wrt}{with respect to~}
\newcommand{\re}{r.e.}
\newcommand{\nind}{\noindent}
\newcommand{\distributed}{distributed\ }
\newcommand{\bn}{\bigskip\markright{NOTES}
\section*{Notes}}
\newcommand{\Exer}{
\bigskip\markright{EXERCISES}
\section*{Exercises}}
\newcommand{\DG}{D_G}
\newcommand{\Sm}{{\rm S5}_m}
\newcommand{\Smc}{{\rm S5C}_m}
\newcommand{\Smi}{{\rm S5I}_m}
\newcommand{\Smic}{{\rm S5CI}_m}
\newcommand{\Martin}{Mart\'\i n\ }
\newcommand{\ol}{\setlength{\itemsep}{0pt}\begin{enumerate}}
\newcommand{\eol}{\end{enumerate}\setlength{\itemsep}{-\parsep}}
\newcommand{\ul}{\setlength{\itemsep}{0pt}\begin{itemize}}
\newcommand{\dl}{\setlength{\itemsep}{0pt}\begin{description}}
\newcommand{\edl}{\end{description}\setlength{\itemsep}{-\parsep}}
\newcommand{\eul}{\end{itemize}\setlength{\itemsep}{-\parsep}}
\newtheorem{fthm}{Theorem}
\newtheorem{flem}[fthm]{Lemma}
\newtheorem{fcor}[fthm]{Corollary}
\newcommand{\slidehead}[1]{
\eject
\Huge
\begin{center}
{\bf #1 }
\end{center}
\vspace{.5in}
\LARGE}
 
\newcommand{\subG}{_G}
\newcommand{\If}{{\bf if}}
 
\newcommand{\attime}{{\tt \ at\_time\ }}
\newcommand{\hatell}{\skew6\hat\ell\,}
\newcommand{\Then}{{\bf then}}
\newcommand{\Until}{{\bf until}}
\newcommand{\Else}{{\bf else}}
\newcommand{\Repeat}{{\bf repeat}}
\newcommand{\cA}{{\cal A}}
\newcommand{\cE}{{\cal E}}
\newcommand{\cF}{{\cal F}}
\newcommand{\cI}{{\cal I}}
\newcommand{\cN}{{\cal N}}
\newcommand{\cR}{{\cal R}}
\newcommand{\cS}{{\cal S}}
\newcommand{\BN}{B^{\scriptscriptstyle \cN}}
\newcommand{\BS}{B^{\scriptscriptstyle \cS}}
\newcommand{\cW}{{\cal W}}
\newcommand{\EG}{E_G}
\newcommand{\CG}{C_G}
\newcommand{\CN}{C_\cN}
\newcommand{\ES}{E_\cS}
\newcommand{\EN}{E_\cN}
\newcommand{\CS}{C_\cS}

\newcommand{\attack}{\mbox{{\it attack}}}
\newcommand{\attacking}{\mbox{{\it attacking}}}
\newcommand{\delivered}{\mbox{{\it delivered}}}
\newcommand{\exist}{\mbox{{\it exist}}}
\newcommand{\decide}{\mbox{{\it decide}}}
\newcommand{\clean}{{\it clean}}
\newcommand{\diff}{{\it diff}}
\newcommand{\Failed}{{\it failed}}
\newcommand\eqdef{=_{\rm def}}
\newcommand{\true}{\mbox{{\it true}}}
\newcommand{\false}{\mbox{{\it false}}}
 
\newcommand{\DN}{D_{\cN}}
\newcommand{\DS}{D_{\cS}}
\newcommand{\tyme}{{\it time}}
\newcommand{\fp}{f}
 
\newcommand{\Kax}{{\rm K}_n}
\newcommand{\Kaxc}{{\rm K}_n^C}
\newcommand{\Kaxd}{{\rm K}_n^D}
\newcommand{\Tax}{{\rm T}_n}
\newcommand{\Taxc}{{\rm T}_n^C}
\newcommand{\Taxd}{{\rm T}_n^D}
\newcommand{\fourax}{{\rm S4}_n}
\newcommand{\fouraxc}{{\rm S4}_n^C}
\newcommand{\fouraxd}{{\rm S4}_n^D}
\newcommand{\fiveax}{{\rm S5}_n}
\newcommand{\fiveaxc}{{\rm S5}_n^C}
\newcommand{\fiveaxd}{{\rm S5}_n^D}
\newcommand{\Dax}{{\rm KD45}_n}
\newcommand{\Daxc}{{\rm KD45}_n^C}
\newcommand{\Daxd}{{\rm KD45}_n^D}
\newcommand{\LP}{{\cal L}_n}
\newcommand{\LCP}{{\cal L}_n^C}
\newcommand{\LDP}{{\cal L}_n^D}
\newcommand{\LCDP}{{\cal L}_n^{CD}}
\newcommand{\MP}{{\cal M}_n}
\newcommand{\MPr}{{\cal M}_n^r}
\newcommand{\MPrt}{\M_n^{\mbox{\scriptsize{{\it rt}}}}}
\newcommand{\MPrst}{\M_n^{\mbox{\scriptsize{{\it rst}}}}}
\newcommand{\MPelt}{\M_n^{\mbox{\scriptsize{{\it elt}}}}}
\renewcommand{\lang}{\mbox{${\cal L}_{n} (\Phi)$}}
\renewcommand{\langd}{\mbox{${\cal L}_{n}^D (\Phi)$}}
\newcommand{\fiveaxdu}{{\rm S5}_n^{DU}}
\newcommand{\LPD}{{\cal L}_n^D}
\newcommand{\fiveaxu}{{\rm S5}_n^U}
\newcommand{\fiveaxcu}{{\rm S5}_n^{CU}}
\newcommand{\LPU}{{\cal L}^{U}_n}
\newcommand{\LPCU}{{\cal L}_n^{CU}}
\newcommand{\LDPU}{{\cal L}_n^{DU}}
\newcommand{\LCPU}{{\cal L}_n^{CU}}
\newcommand{\LPDU}{{\cal L}_n^{DU}}
\newcommand{\LPCDU}{{\cal L}_n^{\it CDU}}
\newcommand{\Cn}{\C_n}
\newcommand{\CSnp}{\I_n^{oa}(\Phi')}
\newcommand{\CSc}{\C_n^{oa}(\Phi)}
\newcommand{\Ccs}{\C_n^{oa}}
\newcommand{\CSAX}{OA$_{n,\Phi}$}
\newcommand{\CSAXN}{OA$_{n,{\Phi}}'$}
\newcommand{\untill}{U}
\newcommand{\until}{\, U \,}
\newcommand{\amp}{{\rm a.m.p.}}
\newcommand{\commentout}[1]{}
\newcommand{\msgc}[1]{ @ #1 }
\newcommand{\Camp}{{\C_n^{\it amp}}}
\newcommand{\bi}{\begin{itemize}}
\newcommand{\ei}{\end{itemize}}
\newcommand{\be}{\begin{enumerate}}
\newcommand{\ee}{\end{enumerate}}
\newcommand{\rarrowr}{\stackrel{r}{\rightarrow}}
\newcommand{\ack}{\mbox{\it ack}}
\newcommand{\Gz}{\G_0}
\newcommand{\denselist}{\itemsep 0pt\partopsep 0pt}
\def\seealso#1#2{({\em see also\/} #1), #2}

\setlength{\evensidemargin}{0in}
\setlength{\oddsidemargin}{0in}
\setlength{\textwidth}{6.25in}
\setlength{\textheight}{8.5in}
\setlength{\topmargin}{0in}
\setlength{\headheight}{0in}
\setlength{\headsep}{0in}
\setlength{\itemsep}{0pt}
\renewcommand{\topfraction}{.9}
\renewcommand{\textfraction}{.1}
\setlength{\parskip}{\smallskipamount}

\newenvironment{oldthm}[1]{\par\noindent{\bf Theorem #1:} \em \noindent}{\par}
\newenvironment{oldlem}[1]{\par\noindent{\bf Lemma #1:} \em \noindent}{\par}
\newenvironment{oldcor}[1]{\par\noindent{\bf Corollary #1:} \em \noindent}{\par}
\newenvironment{oldpro}[1]{\par\noindent{\bf Proposition #1:} \em \noindent}{\par}
\newcommand{\othm}[1]{\begin{oldthm}{\ref{#1}}}
\newcommand{\eothm}{\end{oldthm} \medskip}
\newcommand{\olem}[1]{\begin{oldlem}{\ref{#1}}}
\newcommand{\eolem}{\end{oldlem} \medskip}
\newcommand{\ocor}[1]{\begin{oldcor}{\ref{#1}}}
\newcommand{\eocor}{\end{oldcor} \medskip}
\newcommand{\opro}[1]{\begin{oldpro}{\ref{#1}}}
\newcommand{\eopro}{\end{oldpro} \medskip}

\newcommand{\world}{W}
\newcommand{\WN}{\W_N}
\newcommand{\Winf}{\W^*}
\newcommand{\tends}{\rightarrow}
\newcommand{\tendsto}{\tends}
\newcommand{\ninfty}{{N \rightarrow \infty}}
\newcommand{\nworldsv}[1]{{\it \#worlds}^{#1}}
\newcommand{\nworlds}{{{\it \#worlds}}_{N}^{\epsvec}}
\newcommand{\nwrldPnt}[1]{\nworlds[#1]}
\newcommand{\nworldsarg}[1]{\nworlds[#1]}
\newcommand{\binco}[2]{{{#1}\choose{#2}}}
\newcommand{\closure}[1]{{\overline{#1}}}
\newcommand{\balpha}{\bar{\alpha}}
\newcommand{\bbeta}{\bar{\beta}}
\newcommand{\bgamma}{\bar{\gamma}}
\newcommand{\half}{\frac{1}{2}}
\newcommand{\bQ}{\overline{Q}}
\newcommand{\Vector}[1]{{\langle #1 \rangle}}
\newcommand{\Algzeroone}{\mbox{\em Compute01}}
\newcommand{\Algcompute}{\mbox{{\em Compute-Pr}$_\infty$}}

\newcommand{\Prinfv}[1]{{\Pr}^{#1}_\infty}
\newcommand{\PrNv}[1]{{\Pr}^{#1}_N}
\newcommand{\Prinf}{{\Pr}_\infty}
\newcommand{\PrN}{{\Pr}_N}
\newcommand{\pN}{\PrN (\phi | \KB)}
\newcommand{\IPrinf}{{\Box\Diamond\Prinf}}
\newcommand{\PPrinf}{{\Diamond\Box\Prinf}}
\newcommand{\prNw}{{\Pr}_{N}^{\epsvec}}
\newcommand{\prNwi}{{\Pr}_{N^i}^{\epsvec^i}}
\newcommand{\priw}{{\Pr}_{\infty}}
\newcommand{\beliefprob}[1]{\Pr(#1)} 
\newcommand{\Prinfeps}{{{\Pr}_\infty^{\epsvec}}}
\newcommand{\PrNeps}{{{\Pr}_N^{\epsvec}}}
\newcommand{\Pinf}[2]{{\Pi_\infty^{#1}[#2]}}

\newcommand{\infvocab}{\Omega}
\newcommand{\infunvocab}{\Upsilon}
\newcommand{\vocab}{\Phi}
\newcommand{\nonunaryvocab}{\vocab}
\newcommand{\unvocab}{\Psi}

\newcommand{\cL}{{\cal L}}
\newcommand{\cLne}{{\cal L}^-}
\newcommand{\finitelang}{\cL_i^d(\Phi)}
\newcommand{\fLd}{\cL_d^d(\Phi)}
\newcommand{\cLd}{\cL_{\mbox{\em \scriptsize$d$}}(\Phi)}
\newcommand{\Laeq}{{\cal L}^{\aeq}}
\newcommand{\Leq}{{\cal L}^{=}}
\newcommand{\Lunaeq}{\Laeq_1}
\newcommand{\Luneq}{\Leq_1}

\newcommand{\KB}{{\it KB}}
\newcommand{\phieven}{\phi_{\mbox{\footnotesize\it even}}}
\newcommand{\phiodd}{\phi_{\mbox{\footnotesize\it odd}}}
\newcommand{\KBM}{\KB_{\boldM}}
\newcommand{\Pstar}{P^*}
\newcommand{\barP}{\neg P}
\newcommand{\barQ}{\neg Q}
\newcommand{\maxarity}{\rho}
\newcommand{\Init}{\mbox{\it Init\/}}
\newcommand{\Rep}{\mbox{\it Rep\/}}
\newcommand{\Acc}{\mbox{\it Acc\/}}
\newcommand{\Comp}{\mbox{\it Comp\/}}
\newcommand{\Step}{\mbox{\it Step\/}}
\newcommand{\Univ}{\mbox{\it Univ\/}}
\newcommand{\Exis}{\mbox{\it Exis\/}}
\newcommand{\Between}{\mbox{\it Between\/}}
\newcommand{\Count}{\mbox{\it count\/}}
\newcommand{\phiq}{\phi_Q}
\newcommand{\propform}{\beta}
\newcommand{\allphi}{\ast}
\newcommand{\ID}{\mbox{\it ID}}
\newcommand{\atomdesc}{{\psi_*}}
\newcommand{\rigid}{\mbox{\it rigid}}
\newcommand{\abdesc}{\widehat{D}}
\newcommand{\KBtwo}{\theta}
\newcommand{\psits}{\psi[\KBtwo,\abdesc]}
\newcommand{\psitsnohat}{\psi[\KBtwo,D]}
\newcommand{\KBfly}{\KB_{\mbox{\scriptsize \it fly}}}
\newcommand{\KBchirps}{\KB_{\mbox{\scriptsize \it chirps}}}
\newcommand{\KBmagpie}{\KB_{\mbox{\scriptsize \it magpie}}}
\newcommand{\KBhep}{\KB_{\mbox{\scriptsize \it hep}}}
\newcommand{\KBnixon}{\KB_{\mbox{\scriptsize \it Nixon}}}
\newcommand{\KBel}{\KB_{\mbox{\scriptsize \it likes}}}
\newcommand{\KBtax}{\KB_{\mbox{\scriptsize \it taxonomy}}}
\newcommand{\KBarm}{\KB_{\mbox{\scriptsize \it arm}}}
\newcommand{\KBlate}{\KB_{\mbox{\scriptsize \it late}}}
\newcommand{\KBp}{{\KB'}}
\newcommand{\KBflyp}{\KB_{\mbox{\scriptsize \it fly}}'}
\newcommand{\KBpp}{{\KB''}}
\newcommand{\KBdef}{\KB_{\mbox{\scriptsize \it def}}}
\newcommand{\KBdishep}{\KB_{\mbox{\scriptsize \it $\lor$hep}}}

\newcommand{\canKB}{\widehat{\KB}}
\newcommand{\canxi}{\widehat{\xi}}
\newcommand{\KBfo}{\KB_{\it fo}}
\newcommand{\KBconst}{\psi}
\newcommand{\KBprop}{\KBp}

\newcommand{\quak}{\mbox{\it Quaker\/}}
\newcommand{\repub}{\mbox{\it Republican\/}}
\newcommand{\pac}{\mbox{\it Pacifist\/}}
\newcommand{\Nixon}{\mbox{\it Nixon\/}}
\newcommand{\Winner}{\mbox{\it Winner\/}}
\newcommand{\Child}{\mbox{\it Child\/}}
\newcommand{\Boy}{\mbox{\it Boy\/}}
\newcommand{\Tall}{\mbox{\it Tall\/}}
\newcommand{\Elephant}{\mbox{\it Elephant\/}}
\newcommand{\Gray}{\mbox{\it Gray\/}}
\newcommand{\Yellow}{\mbox{\it Yellow\/}}
\newcommand{\Clyde}{\mbox{\it Clyde\/}}
\newcommand{\Tweety}{\mbox{\it Tweety\/}}
\newcommand{\Opus}{\mbox{\it Opus\/}}
\newcommand{\Bird}{\mbox{\it Bird\/}}
\newcommand{\Penguin}{{\it Penguin\/}}
\newcommand{\Fish}{\mbox{\it Fish\/}}
\newcommand{\Fly}{\mbox{\it Fly\/}}
\newcommand{\Warmblooded}{\mbox{\it Warm-blooded\/}}
\newcommand{\White}{\mbox{\it White\/}}
\newcommand{\Red}{\mbox{\it Red\/}}
\newcommand{\Winged}{\mbox{\it Winged\/}}
\newcommand{\Giraffe}{\mbox{\it Giraffe\/}}
\newcommand{\Visible}{\mbox{\it Easy-to-see\/}}
\newcommand{\Bat}{\mbox{\it Bat\/}}
\newcommand{\Blue}{\mbox{\it Blue\/}}
\newcommand{\Fever}{\mbox{\it Fever\/}}
\newcommand{\Jaun}{\mbox{\it Jaun\/}}
\newcommand{\Hep}{\mbox{\it Hep\/}}
\newcommand{\Eric}{\mbox{\it Eric\/}}
\newcommand{\Alice}{\mbox{\it Alice\/}}
\newcommand{\Tom}{\mbox{\it Tom\/}}
\newcommand{\Lottery}{\mbox{\it Lottery\/}}
\newcommand{\Zookeeper}{\mbox{\it Zookeeper\/}}
\newcommand{\Fred}{\mbox{\it Fred\/}}
\newcommand{\Likes}{\mbox{\it Likes\/}}
\newcommand{\Day}{\mbox{\it Day\/}}
\newcommand{\Nextday}{\mbox{\it Next-day\/}}
\newcommand{\Sleepslate}{\mbox{\it To-bed-late\/}}
\newcommand{\Riseslate}{\mbox{\it Rises-late\/}}
\newcommand{\TS}{\mbox{\it TS\/}}
\newcommand{\EEJ}{\mbox{\it EEJ\/}}
\newcommand{\FC}{\mbox{\it FC\/}}
\newcommand{\Dodo}{\mbox{\it Dodo\/}}
\newcommand{\Ab}{\mbox{\it Ab\/}}
\newcommand{\Chirps}{\mbox{\it Chirps\/}}
\newcommand{\Swims}{\mbox{\it Swims\/}}
\newcommand{\Magpie}{\mbox{\it Magpie\/}}
\newcommand{\Moody}{\mbox{\it Moody\/}}
\newcommand{\SomeMorning}{\mbox{\it Tomorrow\/}}
\newcommand{\Animal}{\mbox{\it Animal\/}}
\newcommand{\Sparrow}{\mbox{\it Sparrow\/}}
\newcommand{\Turtle}{\mbox{\it Turtle\/}}
\newcommand{\Older}{\mbox{\it Over60\/}}
\newcommand{\Patient}{\mbox{\it Patient\/}}
\newcommand{\Black}{\mbox{\it Black\/}}
\newcommand{\Ray}{\mbox{\it Ray\/}}
\newcommand{\Reiter}{\mbox{\it Reiter\/}}
\newcommand{\Drew}{\mbox{\it Drew\/}}
\newcommand{\McDermott}{\mbox{\it McDermott\/}}
\newcommand{\Emu}{\mbox{\it Emu\/}}
\newcommand{\Canary}{\mbox{\it Canary\/}}
\newcommand{\BlueCanary}{\mbox{\it BlueCanary\/}}
\newcommand{\FlyingBird}{\mbox{\it FlyingBird\/}}
\newcommand{\UL}{\mbox{\it LeftUsable\/}}
\newcommand{\UR}{\mbox{\it RightUsable\/}}
\newcommand{\BL}{\mbox{\it LeftBroken\/}}
\newcommand{\BR}{\mbox{\it RightBroken\/}}
\newcommand{\Ticket}{\mbox{\it Ticket\/}}
\newcommand{\BlueEyed}{{\mbox{\it BlueEyed\/}}}
\newcommand{\Jaundice}{{\it Jaundice\/}}
\newcommand{\Hepatitis}{{\it Hepatitis\/}}
\newcommand{\HeartDisease}{{\mbox{\it Heart-disease\/}}}
\newcommand{\bJ}{{\overline{J}\,}}
\newcommand{\bH}{{\overline{H}\,}}
\newcommand{\bB}{{\overline{B}\,}}
\newcommand{\Prem}{{\mbox{\it Child\/}}}
\newcommand{\David}{{\mbox{\it David\/}}}
\newcommand{\Son}{{\mbox{\it Son\/}}}	

\newcommand{\ceslim}{Ces\`{a}ro limit}
\newcommand{\Sigmad}{\Sigma^d_i}
\newcommand{\Liogonkii}{Liogon'ki\u\i}
\newcommand{\Vstar}{{\modfrag_*}}
\newcommand{\moddesc}{\psi \land \modfrag}
\newcommand{\sumact}{\Degr_2}
\newcommand{\degree}{\Degr_1}
\newcommand{\Active}{\alpha}
\newcommand{\ActiveAtoms}{\boldA}
\newcommand{\named}{n}
\newcommand{\aactive}{a}
\newcommand{\degr}{\delta}
\newcommand{\chsize}{f}
\newcommand{\chconst}{g}
\newcommand{\Chconst}{G}
\newcommand{\Degr}{\Delta}
\newcommand{\frags}{\M}
\newcommand{\const}{H}
\newcommand{\modfrag}{\V}
\newcommand{\AD}{\A}
\newcommand{\Named}{\nu}
\newcommand{\bit}{b}
\newcommand{\guess}{\gamma}
\newcommand{\bxor}[1]{\dot{\bor}}
\newcommand{\weight}{\omega}
\newcommand{\arity}{{\it arity}}
\newcommand{\assigned}{\leftarrow}
\newcommand{\snum}[2]{{{#1} \brace {#2}}}

\newcommand{\eps}{\tau}
\newcommand{\vareps}{\varepsilon}
\newcommand{\varepsvec}{{\vec{\vareps}}}
\newcommand{\xtuple}{\vec{x}}
\newcommand{\ctuple}{{\vec{c}\,}}
\newcommand{\uvec}{{\!{\vec{\,u}}}}
\newcommand{\ucom}{u}
\newcommand{\pvec}{{\!{\vec{\,p}}}}
\newcommand{\pcom}{p}
\newcommand{\vvec}{\vec{v}}
\newcommand{\wvec}{\vec{w}}
\newcommand{\xvec}{\vec{x}}
\newcommand{\yvec}{\vec{y}}
\newcommand{\zvec}{\vec{z}}
\newcommand{\zerovec}{\vec{0}}

\newcommand{\epscom}{\eps}
\newcommand{\epsvec}{{\vec{\eps}\/}}
\newcommand{\prop}[2]{{||{#1}||_{{#2}}}}
\newcommand{\aeq}{\approx} 
\newcommand{\app}{\approx}
\newcommand{\alt}{\prec}
\newcommand{\aleq}{\preceq}
\newcommand{\agt}{\succ}
\newcommand{\ageq}{\succeq}
\newcommand{\altne}{\prec}
\newcommand{\naeq}{\not\approx}
\newcommand{\cprop}[3]{{\|{#1}|{#2}\|_{{#3}}}}
\newcommand{\Bigcprop}[3]{{\Bigl\|{#1}\Bigm|{#2}\Bigr\|_{{#3}}}}

\newcommand{\reals}{\IR}
\newcommand{\qsep}{\,}
\newcommand{\perm}{\pi}
\newcommand{\val}{V}
\newcommand{\rwent}{\mbox{$\;|\!\!\!\sim$}_{\mbox{\scriptsize \it rw}}\;}
\newcommand{\notrwent}{\mbox{$\;|\!\!\!\not\sim$}_{\mbox{\scriptsize\it rw}}\;}
\newcommand{\dempster}{\delta}
\newcommand{\dentails}{{\;|\!\!\!\sim\;}}
\newcommand{\notdentails}{{\;|\!\!\!\not\sim\;}}
\newcommand{\dentailssub}[1]{\dentails\hspace{-0.4em}_{
          \mbox{\scriptsize{\it #1}}}\;}
\newcommand{\notdentailssub}[1]{\notdentails\hspace{-0.4em}_{
          \mbox{\scriptsize{\it #1}}}\;}
\newcommand{\default}{\rightarrow}

\newcommand{\vecof}[1]{{\pi({#1})}}
\newcommand{\PIN}[2]{{\Pi_N^{#1}[#2]}}
\newcommand{\SS}[2]{{S^{#1}[#2]}}
\newcommand{\SSc}[2]{{S^{#1}[#2]}}
\newcommand{\SSzero}[1]{\SS{\zerovec}{#1}}
\newcommand{\SSczero}[1]{\SSc{\zerovec}{#1}}
\newcommand{\SSpos}[1]{\SS{\leq \zerovec}{#1}}
\newcommand{\Sol}{{\it Sol}}
\newcommand{\poscon}{\gamma}
\newcommand{\constraints}{\Gamma}
\newcommand{\constpos}{\constraints^{\leq}}
\newcommand{\propspace}{\Delta^K}
\newcommand{\mept}{\vvec}
\newcommand{\mecoord}{v}
\newcommand{\mepts}{{\Q}}
\newcommand{\OS}{{\cal O}}
\renewcommand{\S}{{\cal S}}
\newcommand{\meval}{{\rho}}
\newcommand{\meptmin}{{\uvec^{\ast}_{\mbox{\scriptsize min}}}}
\newcommand{\sizeof}[1]{{\sigma(#1)}}
\newcommand{\mesize}{{\sigma^{\ast}}}
\renewcommand{\pf}{\alpha}
\newcommand{\ps}{\beta}
\newcommand{\Atoms}{\AD}
\newcommand{\limNstar}{{\lim_{N \rightarrow \infty}}^{\!\!\!*}\:}
\newcommand{\probf}[2]{F_{[#1|#2]}}
\newcommand{\probfun}[1]{F_{[#1]}}
\newcommand{\Prmu}[1]{{\mu}_{#1}}
\newcommand{\foversion}[1]{\xi_{#1}}
\newcommand{\fly}{\mbox{\it fly\/}}
\newcommand{\bird}{\mbox{\it bird\/}}
\newcommand{\yellow}{\mbox{\it yellow\/}}
\newcommand{\propconsts}{\Lambda}
\newcommand{\alldiff}{\chi^{\neq}}
\newcommand{\unaryD}{D^1}
\newcommand{\nonunD}{D^{> 1}}
\newcommand{\eqD}{D^{=}}

%
\renewcommand{\L}{{\cal L}}
\renewcommand{\S}{{\cal S}}

\newcommand{\Sys}{\I}

\newcommand{\Next}{\Circ}
\newcommand{\Eventual}{\Diamond}
\newcommand{\Hence}{\Box}
\newcommand{\Cond}{\mbox{\boldmath$\rightarrow$\unboldmath}}
\newcommand{\RCond}{>}
\newcommand{\Condi}{\Cond_i\,}
\newcommand{\RCondi}{\RCond_i\,}
\newcommand{\scap}[1]{\mbox{\bf #1}}
\newcommand{\Know}{K}
\newcommand{\Bel}{B}

\newcommand{\True}{\mbox{\it true}}
\newcommand{\False}{\mbox{\it false}}
\renewcommand{\Cn}{\mathop{\mbox{Cn}}}
\newcommand{\intension}[1]{[\![ #1 ]\!]}

\newcommand{\pl}{\mbox{\rm pl\/}}
\newcommand{\Pl}{\mbox{\rm Pl\/}}
\newcommand{\PL}{\mbox{\it PL}}
\newcommand{\PBox}{N}
\newcommand{\bottom}{\perp}
\newcommand{\sPL}{{\mbox{\scriptsize\it PL}}}
\newcommand{\Fmapp}{\mathbin{\mbox{$\bigcirc\mskip-16mu+$}}}
\newcommand{\Gmapp}{\mathbin{\mbox{$\bigcirc\mskip-16mu\times$}}}
\newcommand{\Hmapp}{\mathbin{\mbox{$\bigcirc\mskip-16mu\div$}}}

\newcommand{\Poss}{\mbox{Poss}}
\newcommand{\sPoss}{\mbox{\scriptsize\it Poss}}

\newcommand{\Ind}{\mbox{\it Ind}}
\newcommand{\IND}{\mbox{\it IND}}

\newcommand{\LCond}{\L^{C}}
\newcommand{\SysP}{system~{\bf P}}

\newcommand{\Union}{\bigcup}

\newcommand{\Tref}[1]{Theorem~\ref{#1}}
\newcommand{\Lref}[1]{Lemma~\ref{#1}}
\newcommand{\Pref}[1]{Proposition~\ref{#1}}
\newcommand{\Cref}[1]{Corollary~\ref{#1}}
\newcommand{\Chref}[1]{Chapter~\ref{#1}}
\newcommand{\Xref}[1]{Example~\ref{#1}}
\newcommand{\Eref}[1]{Eq.~(\ref{#1})}
\newcommand{\Sref}[1]{Section~\ref{#1}}
\newcommand{\BEL}{\mbox{Bel}}

\newcommand{\Sub}{\mbox{{\em Sub}}}
\newcommand{\Subp}{\mbox{{\em Sub}}^+}
\newcommand{\Subc}{\mbox{{\em Sub}}_C}
\newcommand{\sSub}{\mbox{{\small {\em Sub}}}}
\newcommand{\sSubc}{\mbox{{\small {\em Sub}}}_C}

\newcommand{\Card}[1]{\left| #1\right|}


\newenvironment{RETHM}[2]{\it \trivlist \item[\hskip \labelsep{\bf #1 \ref{#2}:}]}{\endtrivlist}
\newcommand{\rethm}[1]{\begin{RETHM}{Theorem}{#1}}
\newcommand{\repro}[1]{\begin{RETHM}{Proposition}{#1}}
\newcommand{\relem}[1]{\begin{RETHM}{Lemma}{#1}}
\newcommand{\recor}[1]{\begin{RETHM}{Corollary}{#1}}

\newcommand{\erethm}{\end{RETHM}}
\newcommand{\erepro}{\end{RETHM}}
\newcommand{\erelem}{\end{RETHM}}
\newcommand{\erecor}{\end{RETHM}}

\newcommand{\SCond}[3]{#2 \leadsto_{#1} #3}
\renewcommand{\Omega}{W}
\newcommand{\nmon}{{\;|\!\!\!\sim\,}}
\renewcommand{\Cn}{\mathop{\mbox{Cn}}}
\renewcommand{\L}{{\cal L}}
\newcommand{\LCondfo}{{\cal L}^{subj}}
\newcommand{\LCondsfo}{{\cal L}^{stat}}
\newcommand{\LSCondfo}{{\cal L}^{stat}}
\newcommand{\SysCfo}{{\bf C}^{subj}}
\newcommand{\SysCsfo}{{\bf C}^{stat}}
\newcommand{\vdashp}{\vdash_{\mbox{\scriptsize\bf P}}}
\newcommand{\vdashpp}{\vdash_{\mbox{\scriptsize\bf P}$'$}}
\newcommand{\satp}{\sat_{\mbox{\scriptsize p}}}
\newcommand{\satr}{\sat_{\mbox{\scriptsize r}}}
\newcommand{\sate}{\sat_\epsilon}
\newcommand{\Pkappafo}{{\P^\kappa_{subj}}}
\newcommand{\Ppossfo}{{\P^{poss}_{subj}}}
\newcommand{\Ppreffo}{{\P^{p}_{subj}}}
\newcommand{\Pwpreffo}{{\P^{p,w}_{subj}}}
\newcommand{\Pspreffo}{{\P^{p,s}_{subj}}}
\newcommand{\Pepsfo}{{\P^\epsilon_{subj}}}
\newcommand{\Pkappasfo}{{\P^\kappa_{stat}}}
\newcommand{\Pposssfo}{{\P^{poss}_{stat}}}
\newcommand{\Pprefsfo}{{\P^{p}_{stat}}}
\newcommand{\Pwprefsfo}{{\P^{p,w}_{stat}}}
\newcommand{\Psprefsfo}{{\P^{p,s}_{stat}}}
\newcommand{\Pepssfo}{{\P^\epsilon_{stat}}}
\newcommand{\Peps}{{\P^\epsilon}}
\newcommand{\John}{\mbox{\it John\/}}
\newcommand{\Pet}{\mbox{\it Pet\/}}
\newcommand{\Dog}{\mbox{\it Dog\/}}
\newcommand{\Cat}{\mbox{\it Cat\/}}
\newcommand{\Snake}{\mbox{\it Snake\/}}
\newcommand{\homogeneous}{\mbox{\it Homogeneous\/}}
\newcommand{\crooked}{\mbox{\it Crooked\/}}
\newcommand{\lottery}{\mbox{\it Lottery\/}}
\newcommand{\Dec}{\mbox{\it Dec\/}}
\newcommand{\Athlete}{\mbox{\it Athlete\/}}
\newcommand{\Admires}{\mbox{\it Admires\/}}
\newcommand{\Arctic}{\mbox{\it Arctic\/}}
\newcommand{\Married}{\mbox{\it Married\/}}
\newcommand{\Insulated}{\mbox{\it Insulated\/}}
\newcommand{\Nice}{\mbox{\it Nice\/}}
\newcommand{\Pmfo}{\P^m_{fo}}
\newcommand{\PQPLfo}{\P^{QPL}_{subj}}
\newcommand{\PQPLsfo}{\P^{QPL}_{stat}}
\newcommand{\POSS}{\mbox{\it POSS}}
\newcommand{\PREF}{\mbox{\it PREF}}
\newcommand{\PLlot}{\PL_{\mbox{\scriptsize \it lot}}}
\newcommand{\Pllot}{\Pl_{\mbox{\scriptsize \it lot}}}
\newcommand{\vdashL}{\vdash_{\L}}
\newcommand{\setx}{X}
\newcommand{\sety}{Y}
\newcommand{\Dom}{\mbox{\em Dom}}
\newcommand{\Domlot}{\Dom_{\mbox{\scriptsize \it lot}}}
\newcommand{\pilot}{\pi_{\mbox{\scriptsize \it lot}}}
\newcommand{\Wlot}{W_{\mbox{\scriptsize \it lot}}}
\newcommand{\PlDom}{\D}

\begin{titlepage}
\thispagestyle{empty}
\title{
First-Order Conditional Logic Revisited%
\thanks{A preliminary version of this
paper appears in {\em Proc. National Conference on Artificial
Intelligence (AAAI '96)}, 1996, pp.~1305--1312.
Some of this work was done while all three authors were at the
IBM Almaden Research Center, supported by the Air Force Office of Scientific
Research (AFSC) under Contract F49620-91-C-0080; some was done while
Daphne Koller was at U.C.\ Berkeley, supported by a University of California
President's Postdoctoral Fellowship;
and some was done while Nir Friedman
was at Stanford University.  This work was also partially supported by
NSF grants IRI-95-03109 and IRI-96-25901.
}}
\author{
Nir Friedman\\
\small Institute of Computer Science\\
\small Hebrew University\\
\small Jerusalem, 91904 Israel\\
\small nir@cs.huji.ac.il
\and
Joseph Y.\ Halpern\\
\small Dept. of Computer Science\\
\small Cornell University\\
\small Ithaca, NY 14853\\
\small halpern@cs.cornell.edu\\
\and
Daphne Koller\\
\small Dept. of Computer Science\\
\small Stanford University\\
\small Stanford, CA 94305-9010\\
\small koller@cs.stanford.edu
}
\maketitle
\thispagestyle{empty}
\begin{abstract}
{\em  Conditional logics\/} play an important role in recent attempts to
formulate theories of default reasoning. This paper investigates
first-order conditional logic.  We show that, as for first-order
probabilistic logic, it is important not to confound {\em  statistical\/}
conditionals over the domain (such as ``most birds
fly''), and {\em subjective\/} conditionals over possible worlds
(such as ``I believe that Tweety is unlikely to
fly'').  We then address the issue of ascribing semantics to first-order
conditional logic.  As in the propositional case, there are many possible
semantics.   To study the problem in a coherent way, we use
{\em  plausibility structures\/}. These provide us with a general framework in
which many of the standard approaches can be embedded. We show that
while these standard approaches are all the same at the
propositional level, they are significantly different in the context
of a first-order language.
Furthermore, we show that plausibilities provide the
most natural extension of conditional logic to the first-order case:
We provide a sound and complete axiomatization that contains only the
KLM properties and standard axioms of first-order modal logic.  We
show that most of the other approaches have additional properties,
which result in an inappropriate treatment of an infinitary version of
the {\em lottery paradox\/}.
\end{abstract}
\end{titlepage}

\section{Introduction}\label{introduction}
In recent years, conditional logic has
come to play a major role as an
underlying foundation for default reasoning.
Two proposals that have received a lot of attention
\cite{Geffner:thesis,GMPfull} are based on conditional logic.
Unfortunately, while it has long been recognized that first-order expressive
power is necessary for a default reasoning system, most of the work on
conditional logic has been restricted to the propositional case.  In this
paper, we investigate the syntax and semantics of {\em first-order
conditional logic}, with the ultimate goal of providing a first-order
default reasoning system.

Many seemingly different approaches have been proposed for giving
semantics to conditional logic, including preferential structures
\cite{Lewis73,Boutilier94AIJ1,KLM},
$\epsilon$-semantics~\cite{Adams:75,Pearl90},
{\em possibility theory\/}~\cite{DuboisPrade:Defaults91},
and {\em $\kappa$-rankings\/}~\cite{spohn:88,Goldszmidt92}.
In preferential structures, for example, a model consists of a set of
possible worlds, ordered by a preference ordering $\prec$.  If $w \prec w'$,
then the world $w$ is strictly more preferred/more normal than $w'$.  The
formula $\Bird \Cond \Fly$ holds if in the most preferred worlds
in which $\Bird$ holds, $\Fly$ also holds.  (See
Section~\ref{prop-logic}
for more details about this and the other approaches.)

The extension of these approaches to the first-order case seems deceptively
easy.  After all, we can simply have a
preference ordering on first-order,
rather than propositional, worlds.  However, there is a subtlety here.  As
in the case of first-order {\em probabilistic\/} logic~\cite{Bacchus,Hal4},
there are two distinct ways to define conditionals in the first-order case.
In the probabilistic case, the first corresponds to (objective) statistical
statements, such as ``90\% of birds fly''.  The second corresponds to
subjective
degree of belief statements, such as ``the probability that Tweety
(a particular bird) flies is 0.9''.  The first is captured by putting a
probability distribution over the domain (so that the probability of the set
of flying birds is 0.9 that of the set of birds), while the second is
captured by putting a probability on the set of possible worlds (so that the
probability of the set of worlds where Tweety flies is 0.9 that of the set of
worlds where Tweety is a bird).  The same phenomenon occurs in the case of
first-order conditional logic.  Here, we can have a measure (e.g., a
preference order) over the domain, or a measure over the set of
possible
worlds.  The first would allow us to capture qualitative statistical
statements such as ``most birds fly'', while the second would allow us to
capture subjective beliefs such as ``I believe that the bird Tweety is
likely to fly''.  It is important to have a language that allows us to
distinguish between these two very different statements.
Having distinguished between these two types of conditionals, we can ascribe
semantics to each of them using any one of the standard approaches.

There have been
previous attempts to formalize first-order
conditional logic; some are the  natural extension of some
propositional formalism~\cite{Delgrande:Conditional.Logic,%
Brafman97}, while others
use alternative approaches
\cite{LehmannMagidor90,Schlechta95}.
(See Sections~\ref{other.approaches}, \ref{statcase3}, and
\ref{discussion}.
for a more detailed discussion of the alternative approaches.)
How do we make sense of this plethora of
alternatives? Rather
than investigating  them separately, we use a single common framework
that generalizes
almost
all of them.
This framework uses a notion of uncertainty called a {\em plausibility
measure}, introduced by Friedman and Halpern~\citeyear{FrH7}.
A plausibility
measure associates with set of worlds its {\em plausibility}, which is
just an element in a partially ordered space.  Probability measures are a
subclass of plausibility measures, in which the plausibilities lie in
$[0,1]$, with the standard ordering.
Friedman and Halpern~\citeyear{FrH5Full} show that
the different standard approaches to conditional logic
can all be mapped to plausibility measures,
if we interpret $\Bird \Cond \Fly$ as ``the
set of worlds where $\Bird \land \Fly$ holds has greater plausibility
than that of the set of worlds where $\Bird \land \neg \Fly$ holds''.

The existence of a single unifying framework has already proved to be very
useful in the case of propositional conditional logic.  In particular, it
allowed Friedman and Halpern~\citeyear{FrH5Full} to explain the intriguing
``coincidence'' that all of the different approaches to conditional logic
result in an identical reasoning system, characterized by the {\em KLM
postulates\/} \cite{KLM}. In this paper, we show that plausibility
spaces can
also be used to clarify the semantics of first-order conditional logic.
However,
we
show that, unlike the propositional case, the different approaches lead to
different properties in the first-order case.
Of course, these are properties
that require quantifiers and therefore cannot be
expressed in a propositional language.  We show that, in some sense,
plausibilities provide the most natural extension of conditional logic to
the first-order case.  We provide sound and complete axiomatizations for
both the subjective and statistical variants of first-order conditional
logic
that contain only the KLM
properties and the standard axioms of first-order modal logic.%
\footnote{By way of contrast, there is no (recursively enumerable)
    axiomatization of
of either statistical or subjective first-order probabilistic logic;
the validity problem for these logics is highly
undecidable ($\Pi_1^2$ complete) \cite{AH}.}
Essentially the same axiomatizations are shown to be sound and
complete for the first-order version of $\epsilon$-semantics,
but the other
approaches are shown to satisfy additional properties.

One might think that it is not so bad for a conditional logic to satisfy
additional properties.  After all, there are some properties---such as
indifference to irrelevant information---that we would {\em like\/} to be
able to get.  Unfortunately, the additional properties that we get from
using these approaches are not the ones we want.  The properties we get are
related to the treatment of {\em exceptional individuals}.  This issue is
perhaps best illustrated by the {\em lottery paradox\/}
\cite{Kyburg:Rational.Belief}.%
\footnote{We are referring to Kyburg's original version of the lottery
paradox~\cite{Kyburg:Rational.Belief}, and not to the finitary version
discussed by Poole \citeyear{Poole91}.  As Poole showed, any logic of defaults
that satisfies certain minimal properties---properties which are satisfied
by all the logics we consider---is bound to suffer {from} his version of the
lottery paradox.}
Suppose we believe about a lottery that any particular individual typically
does not win the lottery.  Thus we get
\begin{equation}\label{eq1}
\forall x(\true \Cond \neg \Winner(x)).
\end{equation}
However, we believe that typically someone does win the lottery, that is
\begin{equation}\label{eq2}
\true \Cond \exists x \Winner(x).
\end{equation}
Let $\lottery$ be the conjunction of (\ref{eq1}) and (\ref{eq2}).

Unfortunately, in many of the standard approaches,
such as Delgrande's \citeyear{Delgrande:Conditional.Logic} version of
    first-order preferential structures, {from} (\ref{eq1}) we can conclude
\begin{equation}\label{eq3}
\true \Cond \forall x(\neg \Winner(x)).
\end{equation}
Intuitively, {from} (\ref{eq1}) it follows that in the most preferred
worlds, each individual $d$ does not win the lottery.  Therefore, in the
most preferred worlds, no individual wins.  This is exactly what (\ref{eq3})
says.  Since (\ref{eq2}) says that in the most preferred worlds, some
individual wins, it follows that there are no most preferred worlds, \ie we
have $\true \Cond \false$.  While this may be consistent (as it is in
Delgrande's logic),
it implies that all defaults hold, which is surely not what we want.
Of all the approaches,
only $\epsilon$-semantics and plausibility structures,
both of which are fully axiomatized by the first-order extension of the KLM
axioms, do not suffer from this problem.

It may seem that this problem is perhaps not so serious. After all, how
often do we reason about lotteries?  But, in fact, this problem arises in
many situations which are clearly of the type with which we would like to
deal.  Assume, for example, that we express the default ``birds typically
fly'' as Delgrande does, using the statement
\begin{equation}
\label{eq.flybird}
\forall x(\Bird(x) \Cond \Fly(x)).
\end{equation}
Suppose we also believe that Tweety is a bird that does
not fly.  There are a number of ways we can capture beliefs in
conditional logic.  The most standard \cite{FrH1Full} is to
identify belief in $\phi$ with $\phi$ typically being true, that is,
with $\true \Cond \phi$.  Using this approach,
our knowledge base would contain the statement
$\true \Cond
\Bird(\Tweety) \land \neg \Fly(\Tweety)$%
we could similarly conclude $\true
\Cond \false$.  Again, this is surely not what we want.

Our framework allows us to deal with these problems.
Using
plausibilities,
$\lottery$ does not not imply $\True \Cond
\False$, since (\ref{eq3}) does not follow from (\ref{eq1}).  That is, the
lottery paradox simply does not exist if we use plausibilities.
The flying
bird example
is somewhat more subtle.
If we take Tweety to be a {\em nonrigid designator\/} (so that it might
denote different individuals in different worlds), the two statements are
consistent, and the problem disappears.
If, however, Tweety is a rigid designator, the pair is
inconsistent, as we would expect.%
\footnote{To see this, note that if Tweety is a rigid designator, then
$\Bird(\Tweety)\Cond\Fly(\Tweety)$ is a consequence of
(\ref{eq.flybird}). See the discussion in Section~\ref{foplausibility}
for more details on this point.}
This inconsistency suggests that we
    might not
always
want to use
    (\ref{eq.flybird}) to represent ``birds typically fly''.
After all, the former is a statement about a property believed
to hold of each individual bird, while the latter is a statement
about the class of birds.
As argued
in~\cite{BGHKfull}, defaults often arise from statistical facts about
the domain.  That is, the default ``birds typically fly'' is often a
consequence of the empirical observation that ``almost all birds fly''.
By defining a logic which allows us to express statistical conditional
statements, we provide the user an alternative way of representing such
defaults.  We would, of course, like such statements to impact our beliefs
about individual birds.  In~\cite{BGHKfull}, the same issue was addressed in
the probabilistic context, by presenting an approach for going from
statistical knowledge bases to subjective degrees of belief.  We leave the
problem of providing a similar mechanism for conditional logic to future work.

The rest of this paper is organized as follows.
In Section~\ref{prop-logic}, we review
the various approaches to
conditional logic in the propositional case; we also review the definition
of plausibility measures from~\cite{FrH5Full} and show how they provide a common
framework for these different approaches.
In the next three sections, we focus on first-order subjective
conditional logic.   In Section~\ref{foplausibility}, we
describe the syntax for the language and ascribe semantics to formulas
using plausibility.  In Section~\ref{sec:fo-def},
we provide a sound and complete
axiomatization for first-order subjective conditional assertions.
In Section~\ref{other.approaches}, we discuss the
generalization of the other propositional approaches to the first-order
subjective case, by investigating their behavior with respect to the
lottery paradox.
We also provide a brief comparison to some of the other approaches
suggested in the literature.
In Sections~\ref{statcase1}, \ref{statcase2}, and~\ref{statcase3}, we go
through the same exercise for first-order statistical conditional logic,
describing the syntax and semantics, providing a complete
axiomatization, and comparing to other approaches.  We conclude in
Section~\ref{discussion}
with discussion and some directions for further work.

\section{Propositional conditional logic}\label{prop-logic}

The syntax of propositional conditional  logic is simple. We start
    with a set $\Phi$
    of propositions and close off under the usual propositional
    connectives ($\neg$, $\lor$, $\land$, and $\rimp$,
denoting, negation, disjunction, conjunction and material
implication, respectively)
and the
    conditional connective $\Cond$. That is, if $\phi$ and $\psi$ are
    formulas in the language, so
are $\neg \phi, \phi\lor\psi, \phi\land\psi, \phi\rimp\psi$, and $\phi
\Cond \psi$.

Many semantics have been proposed in the literature for conditionals.
Most
of them involve structures of the form $(W,X,\pi)$, where $W$
    is a set of possible worlds, $\pi(w)$ is a truth assignment to
    primitive propositions, and $X$ is some ``measure'' on $W$ such as
    a preference ordering \cite{Lewis73,KLM}.%
\footnote{We could also consider a more general definition,
in which one associates a different ``measure'' with each
world, as done by Lewis \citeyear{Lewis73}.  It is
straightforward to  extend our definitions to
handle this. Since this issue is orthogonal to the main point
of the paper, we do not discuss it further here.}
We now describe some of the proposals in the literature, and
    then show how they can be generalized.
Given a structure $(W,X,\pi)$,
let $\intension{\phi} \subseteq W$ be
    the set of worlds
    satisfying $\phi$.
\begin{itemize}\denselist

 \item A {\em possibility measure\/} \cite{DuboisPrade88} $\Poss$ is a
    function $\Poss:2^{\Omega}\mapsto [0,1]$ such that $\Poss(W) = 1$,
    $\Poss(\emptyset) = 0$, and $\Poss(A) = \sup_{w \in A}(\Poss(\{w\})$.
A {\em  possibility structure\/} is a tuple $(W,\Poss,\pi)$,
    where $\Poss$ is a possibility measure on $W$.
     It satisfies a conditional $\phi\Cond\psi$ if either
    $\Poss(\intension{\phi}) = 0$ or $\Poss(\intension{\phi\land\psi})
    > \Poss(\intension{\phi\land\neg\psi})$ \cite{DuboisPrade:Defaults91}.
That is, either $\phi$ is impossible, in which case the conditional holds
    vacuously, or $\phi\land\psi$ is more possible than
    $\phi\land\neg\psi$.

 \item  A {\em $\kappa$-ranking\/} (or {\em ordinal ranking\/}) on
    $\Omega$ (as defined by \cite{Goldszmidt92}, based on ideas that
    go back to \cite{spohn:88}) is a function $\kappa: 2^\Omega
    \rightarrow \IN^*$, where $\IN^* = \IN \union \{\infty\}$, such
    that  $\kappa(\Omega) = 0$, $\kappa(\emptyset) = \infty$, and
    $\kappa(A) = \min_{w\in A}(\kappa(\{w\}))$. Intuitively, an
    ordinal ranking assigns a degree of surprise to each subset of
    worlds in $\Omega$, where $0$ means unsurprising and higher
    numbers denote greater surprise.
A {\em $\kappa$-structure\/}  is a tuple $(W,\kappa,\pi)$, where
    $\kappa$ is an ordinal ranking on $W$. It satisfies a conditional
    $\phi\Cond\psi$ if either $\kappa(\intension{\phi}) = \infty$ or
    $\kappa(\intension{\phi\land\psi}) <
    \kappa(\intension{\phi\land\neg\psi})$.

\item
A {\em preference ordering\/} on $W$ is a partial order
    $\prec$ over $W$ \cite{KLM,Shoham87}. Intuitively, $w \prec w'$
    holds if $w$ is {\em  preferred\/} to $w'$. A {\em preferential
    structure\/} is a tuple $(W,\prec,\pi)$,
    where $\prec$ is a partial order on $W$. The intuition
    \cite{Shoham87} is that a preferential structure
    satisfies a conditional $\phi\Cond\psi$ if all the  most
    preferred worlds (\ie the minimal worlds according to $\prec$) in
    $\intension{\phi}$ satisfy $\psi$.  However, there may be
no minimal worlds in $\intension{\phi}$. This can happen
    if $\intension{\phi}$ contains an infinite descending
    sequence $\ldots \prec w_2 \prec w_1$.
What do we
do in these structures?  There are
a number of options: the first is to assume that, there are no
infinite descending sequences, \ie that $\prec$ is {\em
well-founded\/}; this is essentially the assumption made by
Kraus, Lehmann, and Magidor \citeyear{KLM}.%
\footnote{Actually, they make a weaker
assumption, called {\em smoothness\/}, that for each formula $\phi$, there
are minimal  worlds in
$\intension{\phi}$, \ie that $\prec$ is well-founded on the sets
of interest.  All the results we prove
for well-founded preferential structures hold for smooth ones as well.}
A yet more general definition---one that works even if $\prec$ is
not well-founded---is given in
    \cite{Lewis73,Boutilier94AIJ1}.
    Roughly speaking, $\phi \Cond \psi$ is true if, from a certain
    point on, whenever $\phi$ is true, so is $\psi$.
More formally,
\begin{quote}
    $(W,\prec,\pi)$ satisfies  $\phi\Cond\psi$ if,
    for every
world $w_1 \in \intension{\phi}$,
    there is a world $w_2$ such that (a) $w_2 \preceq w_1$
(\ie either $w_2 \prec w_1$ or  $w_2 = w_1$ )
    (b) $w_2 \in \intension{\phi\land\psi}$, and (c) for all
    worlds $w_3 \prec w_2$, we have
     $w_3 \in \intension{\phi \rimp \psi}$ (so any world more
     preferred than $w_2$ that satisfies $\phi$ also satisfies $\psi$).
\end{quote}
It is easy to verify that this definition is equivalent to the
earlier one if $\prec$ is well founded.
\item  A {\em parameterized probability distribution \/} (PPD) on $W$ is a
    sequence $ \{\Pr_i : i \ge 0\}$ of
    probability measures over $W$.
A {\em PPD structure\/} is a tuple $(W,\{ \Pr_i : i \ge 0
    \},\pi)$, where $\{\Pr_i\}$ is PPD over $W$.
    Intuitively, it satisfies a conditional $\phi\Cond\psi$ if the
    conditional probability $\psi$ given $\phi$ goes to $1$ in the
    limit. Formally, $\phi\Cond\psi$ is satisfied if $\lim_{i
    \rightarrow \infty}\Pr_i(\intension{\psi}|\intension{\psi}) = 1$
(where
    $\Pr_i(\intension{\psi}|\intension{\phi})$ is taken to be 1 if
    $\Pr_i(\intension{\phi}) = 0$). PPD structures were introduced in
    \cite{GMPfull} as a reformulation of Pearl's
    {\em $\epsilon$-semantics\/} \cite{Pearl90}.
\end{itemize}
These variants are quite different from each other.
As Friedman and Halpern~\citeyear{FrH5Full} show,
we can provide a uniform framework for all of them
    using the notion of plausibility measures.
A {\em plausibility measure\/} $\Pl$ on $W$ is a function that maps
    subsets of $W$ to elements in some arbitrary partially ordered
set.
We read $\Pl(A)$ as ``the plausibility of set $A$''.  If $\Pl(A) \le
\Pl(B)$, then $B$ is at least as plausible as $A$.
Formally, a {\em plausibility space\/} is a tuple
$S = (\Omega,\F, \Pl)$, where $\Omega$ is a set of worlds, $\F$ is an
    algebra of subsets of $\Omega$ (that is, a set of subsets closed
under union and complementation),
    and $\Pl$ maps the sets in $\F$ to
     some set $D$, partially ordered by a relation $\le$
(so that $\le$ is reflexive, transitive, and anti-symmetric).
To simplify notation, we typically omit the algebra $\F$ from the
description of the plausibility space.
As usual, we define the ordering $<$ by taking $d_1 < d_2$ if $d_1 \le
d_2$ and $d_1 \ne d_2$.
We assume that $D$ is {\em pointed\/}: that is,
    it contains two special elements $\top$ and $\bottom$ such
    that $\bottom \le d \le \top$ for all $d \in D$; we
    further assume that $\Pl(\Omega) = \top$ and $\Pl(\emptyset) =
    \bottom$.
Since we want a set to be
    at least as plausible as any of its subsets, we require:
\begin{itemize}\denselist
 \item[A1.] If $A \subseteq B$, then $\Pl(A) \le \Pl(B)$.
\end{itemize}

Clearly, plausibility spaces generalize probability spaces.
Other approaches to dealing with uncertainty, such as possibility
measures, $\kappa$-rankings, and {\em belief functions} \cite{Shaf},
are also easily seen to be plausibility measures.
We can give semantics to conditionals using plausibility in much the same
way as it is done using possibility.
A {\em  plausibility structure\/} is a tuple $\PL = (W,
$\Pl$, \pi)$, where $\Pl$ is a plausibility measure on $W$.
We then define:
\begin{itemize}\denselist
 \item $\PL \sat \phi\Cond\psi$ if either
    $\Pl(\intension{\phi}) = \bottom$ or
    $\Pl(\intension{\phi\land\psi}) >
    \Pl(\intension{\phi\land\neg\psi})$.
\end{itemize}
Intuitively, $\phi \Cond \psi$ holds vacuously if $\phi$ is
    impossible; otherwise, it holds if $\phi \land \psi$ is more
    plausible than $\phi \land \neg \psi$. It is easy to see that this
    semantics for conditionals generalizes the semantics of
    conditionals in
    possibility structures and $\kappa$-structures.
We are implicitly assuming here that $\intension{\phi}$ is
in $\F$ (\ie in the domain of $\Pl$) for each formula $\phi$.
As shown in \cite[Theorem~4.2]{FrH5Full},
    it also generalizes the semantics of conditionals in preferential
    structures and PPD structures.
    More precisely, a mapping is given from
    preferential structures (resp., PPD structures)
to plausibility structures such that the semantics of defaults are
preserved.
For future reference, we sketch these constructions here.

For PPDs, it is quite straightforward.
Given a PPD $PP = (\Pr_1, \Pr_2, \ldots)$ on a space $W$, we can define a
plausibility measure $\Pl_{PP}$ such that $\Pl_{PP}(A) \le \Pl_{PP}(B)$
iff $\lim_{i \tendsto \infty} \Pr_i(B|A \union B) = 1$.  It can then be
shown that $((W,PP,\pi),w) \sat \phi$ iff
$((W,\Pl_{PP},\pi),w) \sat \phi$ for all
$w \in W$ and interpretations $\pi$.

The mapping of preferential structures into plausibility
structures is slightly more complex.
Suppose we are given a preferential structure $(W,\prec,\pi)$.
Let $D_0$ be the domain of
    plausibility values consisting of one element $d_w$ for every
    element $w \in W$.  We use $\prec$ to determine the order of these
    elements: $d_v < d_w$ if $w \prec v$. (Recall that $w \prec w'$
    denotes that $w$ is preferred to $w'$.)  We then take $D$ to be
    the smallest set containing $D_0$ closed under least upper bounds
    (so that every set of elements in $D$ has a least upper bound in
    $D$).
It is not hard to show that $D$ is well-defined (\ie there is a
unique, up to renaming, smallest set) and that taking $\Pl_\prec(A)$ to
be the least upper bound of $\{ d_w : w \in A \}$ gives us the
following property:
\beqn
\parbox{5in}{
$\Pl_\prec(A) \le \Pl_\prec(B)$ if and only if for all $w \in A - B$,
    there is a world $w' \in B$ such that $w' \prec w$ and there is no
    $w'' \in A - B$ such that $w'' \prec w'$.}
\label{eq:prec-embed}
\eeqn
It is then easy to check that $((W,\prec,\pi),w) \sat \phi$ if and
    only if $((W,\Pl_\prec,\pi),w) \sat \phi$,  for all
$w \in W$ and interpretations $\pi$.

These results show that our semantics for conditionals in plausibility
    structures generalizes the various approaches examined in the
    literature. Does it capture our intuitions about conditionals? In
    the AI literature, there has been discussion of the right
    properties of default statements (which are essentially
    conditionals).
While there has been little consensus on what the ``right''
properties for defaults should be, there has been some consensus on
a reasonable ``core'' of inference rules for default reasoning.
This core, is known as the KLM properties \cite{KLM}.
We briefly list these properties here:

\begin{itemize}\denselist
\item[LLE.]
If $\vdash \phi \dimp \phi'$%
\footnote{where $\vdash$ denotes provability in propositional logic},
then from
$\phi\Cond\psi$ infer  $\phi'\Cond\psi$ \hfill (Left Logical
Equivalence)
\item[RW.]
If $\vdash \psi \rimp \psi'$, then from
$\phi\Cond\psi$ infer $\phi\Cond\psi'$ \hfill (Right Weakening)
 \item[REF.] $\phi\Cond\phi$ \hfill (Reflexivity)
 \item[AND.] From $\phi\Cond\psi_1$ and $\phi\Cond\psi_2$ infer
    $\phi\Cond \psi_1 \land \psi_2$ \hfill (And)
 \item[OR.] From $\phi_1\Cond\psi$ and $\phi_2\Cond\psi$ infer
    $\phi_1\lor\phi_2\Cond \psi$  \hfill (Or)
 \item[CM.] From $\phi\Cond\psi_1$ and $\phi\Cond\psi_2$ infer
    $\phi\land \psi_2 \Cond \psi_1$ \hfill (Cautious Monotonicity)
\end{itemize}
LLE states that the syntactic form of
    the antecedent is irrelevant. Thus, if $\phi_1$ and $\phi_2$ are
    equivalent, we can deduce $\phi_2\Cond\psi$ from
    $\phi_1\Cond\psi$.  RW describes a
    similar property of the consequent: If $\psi$ (logically) entails
    $\psi'$, then we can deduce $\phi\Cond\psi'$ from $\phi\Cond\psi$.
    This allows us to can combine default and logical reasoning.
    REF states that $\phi$ is always a default conclusion
    of $\phi$. AND states that we can combine two default conclusions:
    If we can conclude by default both $\psi_1$ and $\psi_2$ from
    $\phi$, then we can also conclude  $\psi_1\land\psi_2$ from $\phi$.
    OR states that we are allowed to reason by cases: If the
    same default conclusion follows from each of two antecedents, then
it also follows from their disjunction. CM
    states that if $\psi_1$ and $\psi_2$ are two default
    conclusions of $\phi$, then discovering that $\psi_2$ holds
   when $\phi$
    holds (as would be
    expected, given the default) should not cause us to retract the
    default conclusion $\psi_1$.

Do conditionals in plausibility structures satisfy  the KLM properties?
They always satisfy REF, LLE, and RW, but they do not in general satisfy
AND, OR, and CM.
To satisfy the KLM properties we must limit
  our attention to plausibility structures that satisfy the following
two conditions:
\begin{itemize}\denselist
 \item[A2.]
If $A$, $B$, and $C$ are pairwise disjoint sets,
 $\Pl(A \union B) > \Pl(C)$, and $\Pl(A \union C) > \Pl(B)$, then
 $\Pl(A) > \Pl(B \union C)$.
 \item[A3.] If $\Pl(A) = \Pl(B) = \bottom$, then $\Pl(A \union B) =
    \bottom$.
\end{itemize}
A plausibility space $(W,\Pl)$ is {\em qualitative\/}
if it satisfies A2 and A3.
A plausibility structure $(W,\Pl,\pi)$ is qualitative
if $(W,\Pl)$ is a qualitative plausibility space.
Friedman and Halpern~\citeyear{FrH5Full} show that, in a very
    general sense, qualitative plausibility structures capture
    default reasoning. More precisely, the KLM properties
    are sound with
    respect to a class of plausibility structures if and only if the
    class consists of qualitative plausibility structures. Furthermore,
    a very weak condition is necessary and sufficient in
    order for the KLM properties to be a complete axiomatization of
conditional logic.  As a consequence,
    once we consider a class
    of structures where the KLM axioms are sound, it is almost
    inevitable that they will also be complete with respect to that class.
This explains the somewhat surprising fact that KLM properties
    characterize default
    entailment not just in preferential structures, but also in
    $\epsilon$-semantics, possibility measures, and $\kappa$-rankings.
    Each one of these approaches
    corresponds, in a precise sense, to a class of qualitative
    plausibility structures. These results show that plausibility
    structures provide a unifying framework for the characterization
    of default entailment in these different logics.

\section{First-order subjective conditional logic}
\label{foplausibility}
We now want to generalize conditional logic to the first-order case.
As mentioned above,
there are two distinct notions of conditionals in
    first-order logic, one involving statistical conditionals and one
involving subjective conditionals.  For each of these, we use a
different syntax,
analogous to the syntax used in \cite{Hal4} for the probabilistic case.
In the next three sections, we focus on the subjective case; in
Sections~\ref{statcase1}, \ref{statcase2}, and~\ref{statcase3}, we
consider the statistical case.

The syntax for subjective conditional logic  is fairly straightforward.
Let $\Phi$ be a first-order vocabulary,
consisting of predicate and function symbols.
(As usual, constant symbols are
viewed as 0-ary function symbols.)
Starting with atomic formulas of first-order logic, we form
more complicated formulas by closing off under
truth-functional connectives (\ie $\land,\lor,\neg$, and
    $\rimp$),
first-order quantification, and the modal operator
$\Cond$.  Thus, a typical formula is
$\forall x (P(x) \Cond \exists y (Q(x,y) \Cond R(y)))$.
Let $\LCondfo(\Phi)$ be the resulting language (the ``subj'' stands for
``subjective'', since the conditionals are viewed as expressing
subjective degrees of belief).
We typically omit the $\Phi$ if it is clear from context or irrelevant.

We can ascribe semantics to subjective conditionals using any one of the
approaches described in the previous section.
However, since we can embed all of the
approaches within the class of plausibility structures, we use these as the
basic semantics.  As in the propositional case, we can then analyze the
behavior of the other approaches simply by restricting attention to the
appropriate subclass of plausibility structures.

To give semantics to $\LCondfo(\Phi)$, we use
{\em (first-order) subjective plausibility structures\/} over $\Phi$.
These are tuples of the form $\PL
    = (\Dom,W, \Pl, \pi)$, where $\Dom$ is a domain,
   $(\Omega,\Pl) $ is a plausibility space and $\pi(w)$ is an
interpretation assigning to each predicate symbol and function symbol in
$\Phi$ a predicate or function of the right arity over $\Dom$.
As usual, a {\em valuation\/} maps each variable to an element of $\Dom$.
We define
the set of worlds that satisfy $\phi$ given the valuation $v$
to be $\intension{\phi}_{({\sPL},v)} =
\{ w\ :\ (\PL,w,v) \sat \phi \}$.
(We omit the subscript whenever it is clear from context.)
For subjective conditionals, we have
\begin{itemize}
\item
$(\PL,w,v) \sat \phi \Cond \psi$ if $\Pl(\intension{\phi}_{({\sPL},v)}) =
    \perp$ or $\Pl(\intension{\phi \land \psi}_{({\sPL},v)}) >
    \Pl(\intension{\phi \land \neg \psi}_{({\sPL},v)})$.
\end{itemize}
The semantics of atomic formulas and quantifiers is the same as in
first-order logic.  As an example, for the atomic formula $P(x,\boldc)$,
we have
\begin{itemize}
\item $(\PL,w,v) \sat P(x,\boldc)$ if $(v(x),\pi(w)(\boldc)) \in
\pi(w)(P)$.
\end{itemize}
Note that $\pi(w)(\boldc)$ is the interpretation of the constant
$\boldc$ in the world $w$.  There may be a different interpretation of
$\boldc$ in each world; that is, we may have $\pi(w)(\boldc) \ne
\pi(w')(\boldc)$ if $w \ne w'$.  Thus, $\boldc$ is {\em nonrigid}.
We return
to this issue below.  Similarly, $\pi(w)(P)$ is the interpretation of
$P$ in $w$.

To give the semantics of quantification, it is useful to define a family
of equivalence relations $\sim_X$ on valuations, where $X$ is a set of
variables.  We write $v \sim_X v'$ if $v$ and $v'$ agree on the values
they give to all variables except possibly those in $X$.  If $X$ is the
singleton $\{x\}$, we write $\sim_x$ instead of $\sim_{\{x\}}$.
\begin{itemize}
\item $(\PL,w,v) \sat \forall x\phi$ if $(\Pl,w,v') \sat \phi$ for
all valuations $v'$ such that $v' \sim_x v$.
\end{itemize}

Because terms are not rigid designators, we cannot substitute
terms for universally quantified variables.  (A similar phenomenon holds
in other modal logics where terms are not rigid \cite{Garson}.)
For example, let $\PBox \phi$ be an abbreviation for
$\neg \phi \Cond \false$.
Notice that $(\PL,w) \sat \PBox \phi$ if $\Pl(\intension{\neg \phi}) =
\bottom$; i.e., $\PBox \phi$ asserts that the plausibility
of $\neg \phi$ is the same as that of the empty set,
so that $\phi$ is true ``almost everywhere''.%
\footnote{$\PBox$ stands for ``necessary''.}
Suppose $\boldc$ is a constant
that does not appear in the formula $\phi$.
It is not hard to see that
$\forall x (\neg \PBox \phi(x)) \rimp (\neg\PBox
\phi(\boldc))$ is not valid in our framework; that is, we cannot
substitute constants for universally quantified variables.
To see this, let $\phi(x)$ be the formula $P(x)$, where $P$ is a unary
predicate.  Consider the plausibility structure
$\PL=(\{d_1,d_2\},\{w_1,w_2\},\Pl,\pi)$, where $\pi$ is such that
$\boldc$ is $d_1$ in world $w_1$ and $d_2$ in world $w_2$, the extension
of $P$ in $w_1$ is $\{d_1\}$ and the extension of $P$ in $w_2$ is
$\{d_2\}$, and $\Pl$
is such that $\Pl(\{w_1\}) = \Pl(\{w_2\} \ne \bottom$.  It is easy to
see that $(\PL,w_1) \sat \forall x (\neg \PBox P(x)) \land \PBox P(c)$.

We could substitute
$\boldc$ for $x$ in $\forall x \phi(x)$
if $\boldc$ were rigid. We can get the effect of
rigidity by assuming that $\exists x (\PBox (x = \boldc))$ holds.  Thus,
we do not lose expressive power by not assuming rigidity.

As in first-order logic, a {\em sentence\/} is a formula with no free
variables.  It is easy to check that, just as in first-order logic, the
truth of a sentence is independent of the valuation.  Thus, if $\phi$ is
a sentence, we often write $(\PL,w) \sat \phi$ rather than $(\PL,w,v)
\sat \phi$.

\section{Axiomatizing first-order subjective conditional logic}
\label{sec:fo-def}
We now want to show that plausibility structures provide an appropriate
semantics for a first-order logic of defaults.
As in the propositional
case, this is true only if we restrict attention to qualitative
plausibility structures, i.e., those satisfying conditions
A2 and A3 above.
Let $\PQPLfo$ be the class of all subjective
qualitative plausibility structures.  We provide a sound and complete
    axiom system for $\PQPLfo$,
and show that it is the natural extension of
the KLM properties
to the first-order case.

The system $\SysCfo$ consists of all generalizations of the following
axioms (where $\phi$ is a {\em generalization\/} of $\psi$ if $\phi$
is of the form $\forall x_1 \ldots \forall x_n \psi$) and rules.
In the axioms
$x$ and $y$ denote variables, while $t$ denotes an arbitrary term.
$\SysCfo$ consists of three parts.
The first set of axioms (C0--C5 together
with the rules MP, R1, and R2) is simply the standard
axiomatization of propositional conditional logic \cite{HC};
the second set (axioms F1--F5) consists of the standard axioms of
first-order logic \cite{Enderton}.
the final set (F6--F7) contains
standard axioms relating the two \cite{HC}.
These axioms describe the interaction between $\PBox$ and
equality, and hold because we are essentially treating variables as
rigid designators.
\begin{itemize}\denselist
 \item[C0.] All instances of propositional tautologies
 \item[C1.] $\phi \Cond \phi$
 \item[C2.]
$((\phi \Cond \psi_1)\land(\phi\Cond\psi_2)) \rimp
    (\phi\Cond(\psi_1\land\psi_2))$
 \item[C3.]
$((\phi_1\Cond\psi)\land(\phi_2\Cond\psi)) \rimp
    ((\phi_1\lor\phi_2) \Cond\psi)$
 \item[C4.]
$((\phi_1\Cond\phi_2)\land(\phi_1\Cond\psi)) \rimp
    ((\phi_1\land\phi_2) \Cond \psi)$
 \item[C5.]
$[(\phi \Cond \psi) \rimp \PBox(\phi \Cond \psi)] \land
            [\neg(\phi \Cond \psi) \rimp \PBox \neg (\phi \Cond \psi)]$
\item[F1.] $\forall x \phi \rimp \phi[x/t]$,
where $t$ is {\em substitutable\/} for $x$ in the sense discussed below
and $\phi[x/t]$ is the result of substituting $t$ for all free
occurrences of $x$ in $\phi$ (see \cite{Enderton} for a formal
definition)
\item[F2.] $\forall x(\phi \rimp \psi) \rimp (\forall x \phi
\rimp \forall x \psi)$
\item[F3.] $\phi \rimp \forall x \phi$ if $x$ does not occur free in
$\phi$
\item[F4.] $x = x$
\item[F5.] $x = y \rimp (\phi \rimp \phi')$, where $\phi$ is a
quantifier-free and $\Cond$-free formula and $\phi'$ is obtained
{f}rom $\phi$ by replacing zero or more occurrences of $x$ in $\phi$
by $y$
\item[F6.] $x = y \rimp \PBox(x=y)$
\item[F7.] $x \ne y \rimp \PBox(x \ne y)$
\item[MP.] From $\phi$ and $\phi \rimp \psi$ infer $\psi$
\item[R1.] From $\phi_1 \dimp \phi_2$
infer $\phi_1\Cond\psi \dimp \phi_2\Cond\psi$
\item[R2.] From $\psi_1 \rimp \psi_2$ infer $\phi\Cond\psi_1 \rimp
    \phi\Cond\psi_2$.
\end{itemize}
It remains
to explain the notion of ``substitutable'' in F1.
Clearly we cannot substitute a term $t$ for $x$ with free variables
that might be captured by some quantifiers in $\phi$; for example,
while $\forall x \exists y (x \ne y)$ is true as long as the domain
has at least two elements, if we substitute $y$ for $x$, we get
$\exists y (y \ne y)$, which is surely false.  In the case of
first-order logic, it suffices to define ``substitutable'' so as
to make sure this does not happen (see
\cite{Enderton} for details).  However, in modal logics such as this one,
we have to be a little more careful. As we observed in
Section~\ref{foplausibility},
we cannot substitute terms for universally
quantified variables in a modal context, since terms are not
in general rigid.
Thus, we
require that if $\phi$ is a formula that has occurrences of $\Cond$, then
the only terms that are substitutable for $x$ in $\phi$
are other variables.

We claim that $\SysCfo$ is the weakest ``natural'' first-order
    extension of the KLM properties.
The bulk of the propositional
fragment of this axiom system (axioms C1--C4, R1, and R2) corresponds
precisely to the KLM properties.
For example, C1 is just REF, C2 is AND, R1 is LLE, and so on.
The remaining axiom (C5)
captures the fact that the plausibility function $\Pl$
is independent of the world.
We could consider a more general semantics where the plausibility
measure used depends on the world (see \cite[Section 8]{FrH5Full}); in this
case, we would drop C5.
This property does not
    appear in \cite{KLM} since they do not allow nesting of
    conditionals. As
    discussed above, the remaining axioms are standard properties of
    first-order modal logic.

The system $\SysCfo$ characterizes first-order default reasoning in
this framework:
\thm\label{subjcompleteness}
$\SysCfo$ is a sound and complete axiomatization of $\LCondfo$
with respect to
$\PQPLfo$. \ethm

\prf The proof combines ideas from the standard Henkin-style
completeness proof for first-order logic \cite{Enderton} with the proof
of completeness for propositional conditional logic given in
\cite{FrH5Full}. The details can be found in the appendix. \eprf

\section{Other approaches to first-order subjective
conditional logic}\label{other.approaches}

In the previous section we showed that $\SysCfo$ is sound and complete
    with respect to $\PQPLfo$.
    What happens if we use one of the approaches
described in Section~\ref{prop-logic} to give semantics to conditionals?
    As noted above,
we can  associate with each of these approach a subset of qualitative
plausibility structures.
 Let
    $\Pwpreffo,\Ppreffo,\Pkappafo,\Ppossfo,$ and $\Pepsfo$ be
    the subsets of $\PQPLfo$ that correspond to
    well-founded preference orderings,
    preference orderings, $\kappa$-rankings, possibility measures,
    and PPDs, respectively.  {F}rom Theorem~\ref{subjcompleteness}, we
immediately get
\thm
$\SysCfo$ is sound in
    $\Pwpreffo$, $\Pspreffo$, $\Ppreffo$, $\Pkappafo$, $\Ppossfo$, and
$\Pepsfo$.
\ethm

Is $\SysCfo$ complete with respect to these approaches?  Even at the
propositional level, it is well known that because $\kappa$ rankings and
possibility measures induce plausibility measures that are total (rather
than partial) orders, they satisfy the following additional property:
\begin{itemize}\denselist
\item[C6.]
$\phi\Cond\psi \land \neg(\phi \Cond \neg\xi) \rimp
(\phi \land \xi \Cond \psi)$.
\end{itemize}
In addition, the plausibility measures induced by $\kappa$ rankings,
possibility measures, and $\epsilon$ semantics are easily seen to have
the property that $\top > \bot$.  This leads to the following axiom:
\begin{itemize}\denselist
\item[C7.] $\neg(\true \Cond \false)$.
\end{itemize}
In the propositional setting, these additional axioms
    and the basic propositional conditional system (\ie C0--C5, MP,
    LLE, and RW) lead to sound and complete axiomatization
    of the corresponding
    (propositional) structures.
(See \cite[Section 8]{FrH5Full}.)

Does the same phenomenon occur in the first-order case?
For $\epsilon$-semantics,
it does.
\thm
\label{sound+complete}
$\SysCfo+${\rm C7} is a sound and complete axiomatization
of $\LCondfo$
with respect to $\Pepsfo$.
\ethm
\prf We combine ideas from the proof of Theorem~\ref{subjcompleteness}
with results from \cite{FrH5Full} showing how a plausibility structure
satisfying C7 can be viewed as a PPD structure.  The details are in the
appendix. \eprf

Although $\epsilon$-semantics has essentially the same expressive power
in the first-order case as plausibility measures, this is not the case
for the other approaches that are characterized by the KLM properties in
the propositional case.
These approaches all satisfy properties beyond $\SysCfo$,
C6, and C7.  And these additional properties are ones that we would argue
are undesirable, since they cause the lottery paradox.
Recall that
$\lottery$, the formula that represents the lottery paradox, is the
conjunction of two formulas:
(\ref{eq1})
$\forall x(\true \Cond \neg \Winner(x))$
states that every individual is unlikely to win the
    lottery, while
(\ref{eq2})
$\true \Cond \exists x \Winner(x)$
states that is is likely that some
    individual does win the lottery.
We start by showing that
$\lottery$ is consistent
    in $\PQPLfo$.

\xam\label{lottery1}
We define a first-order subjective plausibility structure $\PLlot =
    (\Domlot,\Wlot,\Pllot,\pilot)$ as follows:
    $\Domlot$ is a countable domain
    consisting of the individuals $1, 2, 3, \ldots$;
    $\Wlot$ consists of a
    countable number of worlds $w_1, w_2, w_3, \ldots$; $\Pllot$ gives
    the empty set plausibility 0, each non-empty finite set
    plausibility $1/2$, and each infinite set plausibility 1; finally,
    the denotation of $\Winner$ in world $w_i$ according to $\pilot$ is
    the singleton set $\{d_i\}$ (that is, in world $w_i$ the lottery
    winner is individual $d_i$).
It is easy to check that $\intension{\neg \Winner(d_i)} = W -
    \{w_i\}$, so $\Pllot(\intension{\neg \Winner(d_i)}) = 1 > 1/2 =
    \Pl( \intension{\Winner(d_i)})$; hence, $\PLlot$ satisfies
    (\ref{eq1}).  On the other hand, $\intension{\exists x \Winner(x)}
    = W$, so $\Pllot(\intension{\exists x \Winner(x)}) >
    \Pllot(\intension{\neg \exists x \Winner (x)})$; hence $\PLlot$
    satisfies (\ref{eq2}).  It is also easy to verify that $\Pllot$ is
    a qualitative measure, \ie satisfies A2 and A3.  A similar
    construction allows us to capture a situation where birds
    typically fly but we know that Tweety does not fly.
\exam

What happens to the lottery paradox in the other approaches?
First consider
well-founded preferential structures, \ie
    $\Pwpreffo$. In these structures, $\phi \Cond\psi$ holds if $\psi$
    holds in all the preferred worlds that satisfy $\phi$. Thus,
    (\ref{eq1}) implies that for any domain element $d$, $d$ is not a
    winner in the most preferred worlds. On the other hand, (\ref{eq2})
    implies that in the most preferred worlds, some domain element
    wins. Together both imply that there are no preferred worlds.
When, in general, does an argument of this type go through?  As we now show,
it is a consequence of
the following generalization of A2.
\begin{itemize}\denselist
\item[A2$^*$.]
If  $\{ A_i : i \in I \}$ are pairwise disjoint sets, $A =
    \union_{i\in I}A_i$, $0 \in I$, and for all $i \in I - \{0\}$,
    $\Pl(A  - A_i)
    > \Pl(A_i)$, then
    $\Pl(A_0) > \Pl(A - A_0)$.
\end{itemize}
Recall that A2 states that if $A_0$, $A_1$, and $A_2$ are disjoint,
    $\Pl(A_0\union A_1) > \Pl(A_2)$, and
    $\Pl(A_0\union A_2) > \Pl(A_1)$, then $\Pl(A_0) > \Pl(A_1\union
    A_2)$. It is easy to check that for any finite number of
    sets, a similar property follows from A1 and A2 by induction. A2$^*$
asserts that a condition of this type
holds even for an infinite collection of sets.
    This is not
    implied by A1 and A2.
To see this, consider
the plausibility model $\PLlot$
from Example~\ref{lottery1}.
  Take $A_0$ to be empty and take $A_i$, $i >
    1$, to be the singleton consisting of the world $w_i$.  Then
    $\Pllot(A-A_i) = 1 > 1/2 = \Pllot(A_i)$, but $\Pllot(A_0) = 0 < 1
    = \Pl(\union_{i>0} A_i)$.
Hence, A2$^*$ does not hold for plausibility structures in general.  It
does, however, hold for certain subclasses:
\pro\label{A2*}
 A2$^*$ holds in every plausibility structure in $\Pwpreffo$ and $\Pkappafo$.
\epro

\prf See the appendix. \eprf

A consequence of A2$^*$ is the following axiom, called $\forall3$ by
Delgrande:
\begin{itemize}
\item[$\forall 3$.] $\forall x (\phi \Cond \psi) \rimp (\phi \Cond
\forall x \psi)$ if $x$ does not occur free in $\phi$.
\end{itemize}
This axiom
can be viewed as an infinitary version of  axiom C2 (which
is essentially KLM's And Rule), for
(abusing notation somewhat) in a domain $D$, $\forall3$ essentially
says:
$$\land_{d \in D} (\phi \Cond \psi[x/d]) \rimp (\phi \Cond \land_{d \in
 D} \psi[x/d]).$$

\pro\label{forall3} $\forall3$ is valid in all plausibility structures
satisfying A2$^*$. \epro
\prf See the appendix. \eprf

Since A2$^*$ holds in $\Pwpreffo$ and $\Pkappafo$, it follows that
$\forall 3$ does as well.
Moreover,
it is easy to see that the axiom $\forall3$
leads to
the lottery paradox: {From} $\forall x (\true \Cond \neg
\Winner(x))$, $\forall3$
allows us to conclude
$\true \Cond \forall x (\neg
\Winner(x))$.

A2$^*$ does not hold in $\Ppossfo$ and $\Ppreffo$.
In fact, the infinite lottery
is consistent in these classes, although a
somewhat unnatural model is required to express it,
as the following example shows.
\xam Consider
the possibility
structure $(\Domlot,\Wlot,\Poss,\pilot)$, where all the components
besides $\Poss$ are just as in the plausibility
structure $\PLlot$
from Example~\ref{lottery1}
and $\Poss(w_i) = i/(i+1)$.  This means that
if $i > j$, then it is more possible that individual $i$ wins than
individual $j$.  Moreover, this possibility approaches 1 as $i$
increases.  It is not hard to show that this possibility structure
satisfies formulas (\ref{eq1}) and (\ref{eq2}).

A preferential structure in the same spirit also captures the lottery
paradox.  Consider the preferential structure
$(\Domlot,\Wlot,\prec,\pilot)$, where all the components
besides $\prec$ are just as in the plausibility
structure $\PLlot$, and we have $\ldots w_3 \prec w_2 \prec w_2 \prec
w_1$.  Thus, again we have that
if $i > j$, then it is more likely that individual $i$ wins than
individual $j$.  (More precisely, the world where individual $i$ wins is
preferred to that where individual $j$ wins.)  It is easy to verify that
this preferential structure (which is obviously not well-founded) also
satisfies $\lottery$. \exam

Although $\lottery$ is satisfiable in possibility structures and
preferential structures, a slight variant of it is not. Consider a {\em
crooked lottery}, where there is one individual who is
    more likely to win than the rest, but is still unlikely to win.
    To formalize this
    in the language, we add the following formula that we call
    $\crooked$:
$$
\commentout{
\begin{array}{ll}
\neg\exists x(\Winner(x) \Cond\False) \land \exists y\forall x
    (x \neq y \rimp \\
\ \ \     ((\Winner(x) \lor \Winner(y)) \Cond \Winner(y)))
\end{array}
}
\exists y\forall x
(x \neq y \rimp
((\Winner(x) \lor \Winner(y)) \Cond \Winner(y)))
$$
This formula
states that there is an
    individual who is more likely to win than the rest. To see
    this, recall that $(\phi\lor\psi) \Cond \psi$ implies that either
    $\Pl(\intension{\phi\lor\psi}) = \perp$
(which cannot happen here because of the first clause of $\crooked$)
    or $\Pl(\intension{\phi}) < \Pl(\intension{\psi})$.
We take the crooked lottery to be formalized by the formula
$\lottery \land \crooked$.

It is easy to model the crooked lottery using plausibility.
Consider the structure $\PLlot' =
    (\Domlot,\Wlot,\Pllot',\pilot)$, which is identical to $\PLlot$
    except for the plausibility measure $\Pllot'$.
We define $\Pllot'(w_1) = 3/4$; $\Pllot'(w_i) = 1/2$ for $i>1$;
    $\Pllot'(A)$ of a finite set $A$ is $3/4$ if $w_1 \in A$, and
    $1/2$ if $w_1\not\in A$; and $\Pllot(A) = 1$ for infinite $A$.
It is easy to verify that $\PLlot'$ satisfies
    $\crooked$, taking $d_1$ to be the special individual who is
most likely to win (since
    $\Pl(\intension{\Winner(d_1)}) = 3/4 > 1/2 =
    \Pl(\intension{\Winner(d_i)})$ for $i > 1$).  It is also
    easy to verify
    that
$\Pllot' \sat \lottery$.

\commentout{
On the other hand,
the crooked lottery cannot be captured
in $\Ppossfo$ and $\Ppreffo$.

\pro\label{crooked} The formula $\lottery \land \crooked$ is not
satisfiable in $\Ppossfo$ and $\Ppreffo$.
\epro
\prf See the appendix.  \eprf
}
On the other hand, the crooked lottery
cannot be captured in $\Ppossfo$ and $\Ppreffo$.
To show this, we take a slight detour.

Consider the
following two properties:
\begin{itemize}\denselist
\item[A2$^\dagger$.]
If  $\{ A_i : i \in I \}$ are pairwise disjoint sets, $A =
    \union_{i\in I}A_i$, $0 \in I$, and for all $i \in I - \{0\}$,
    $\Pl(A_0) > \Pl(A_i)$, then $\Pl(A_0) \not< \Pl(A - A_0)$.
\item[A3$^*$]
If  $\{ A_i : i \in I \}$ are sets such that $\Pl(A_i) = \bot$, then
$\Pl(\union_i A_i) \bot$.
\end{itemize}

It is easy to see that A2$^\dagger$ is implied by A2$^*$. Suppose
that $\Pl$ satisfies A2$^*$ and the preconditions of A2$^\dagger$. By
A1 we have that $\Pl(A_0) > \Pl(A_i)$ implies that $\Pl(A - A_i) >
\Pl(A_i)$. Thus by A2$^*$ we have that $\Pl(A_0) > \Pl(A - A_0)$, and
therefore $\Pl(A_0) \not < \Pl(A - A_0)$. Moreover, A2$^\dagger$ can
hold in structures that do not satisfy A2$^*$.

\pro\label{A2dagger}
 A2$^\dagger$ holds in every plausibility structure in $\Ppreffo$ and
 $\Ppossfo$.
\epro
\prf See the appendix.  \eprf

A3$^*$ is an infinitary version of A3. It is easy to verify that it
holds in all the approaches we consider, except plausibility measures
and $\epsilon$-semantics.
\pro\label{A3*}
 A3$^*$
 holds in every plausibility structure in $\Ppreffo$,
 $\Pwpreffo$, $\Pkappafo$ and  $\Ppossfo$.
\epro
\prf The proof is straightforward and left as an exercise to the reader. \eprf

A3$^*$ has elegant axiomatic consequences.
\pro\label{A3*axioms}
The axiom $$\forall x \PBox \phi \rimp \PBox(\forall x \phi)$$ is sound
in  structures satisfying A3$^*$.  Moreover, the axiom
$$\forall x (\phi \Cond \psi) \rimp ((\exists x \phi) \rimp \psi),\ \ \mbox{if
$x$ does not appear free in $\psi$}
$$ is sound in structures satisfying A2$^*$ and A3$^*$.%
\footnote{The latter axiom can be viewed as an infinitary version of the
OR Rule (C3), just as $\forall 3$ can be viewed as an infinitary version
of the AND Rule (C2).}
\epro

Finally, we show that when A2$^\dagger$ and A3$^*$ hold, the crooked
lottery is
(almost)
inconsistent.
\pro\label{crooked} The formula $\lottery \land \crooked \rimp (\true
\Cond \false)$ is valid in structures satisfying A2$^\dagger$ and A3$^*$.
\epro
\prf See the appendix.  \eprf
Notice that, since A2$^\dagger$ and A3$^*$ are valid in $\Ppossfo$,
it immediately
follows that $\lottery \land \crooked$ is unsatisfiable in $\Ppossfo$.

To summarize, the discussion in this section shows that, once we move
to first-order logic, kappa-rankings, possibility structures and
preferential structures satisfy extra properties over and above those
characterized by $\SysCfo$ (and C6 and C7). We identified these
properties both in terms of the constraints on the plausibility
measures allowed by these semantics (e.g., conditions A2$^*$,
A2$^\dagger$, and A3$^*$), and in terms of
corresponding properties in the language (e.g., axioms and the
variants of the lottery example). Our analysis leaves open the question of
complete axiomatization of first-order conditional logic with respect
to these classes of structures.

\section{First-order statistical conditional logic}\label{statcase1}
In the next three section, we analyze the statistical version of
first-order
conditional logic in much the same way we did the subjective version.

The syntax for statistical conditionals is fairly straightforward.
Let $\Phi$ be a first-order vocabulary,
consisting of predicate and function symbols.
(As usual, constant symbols are
viewed as 0-ary function symbols.)
Starting with atomic formulas of first-order logic, we form
more complicated formulas by closing off under
truth-functional connectives (\ie $\land,\lor,\neg$, and
    $\rimp$),
first-order quantification,
and the family of modal operators
    $\SCond{\setx}{\phi}{\psi}$,
    where $\setx$ is a set of distinct variables.%
\footnote{This syntax is borrowed from Brafman \citeyear{Brafman97},
which in turn is based on that of \cite{Bacchus,Hal4},
except that in
the earlier papers, the subscript $\setx$ was taken to be a {\em
sequence\/} of variables, rather than a set.  Since the order of the
variables is irrelevant, taking it to be a set seems more natural.}
    We denote the
    resulting language $\LSCondfo(\Phi)$.
(We typically omit the $\Phi$ if it is clear from context.)
The intuitive
    reading of $\SCond{\setx}{\phi}{\psi}$ is that
almost all of the $\setx$'s
that satisfy $\phi$ also satisfy
    $\psi$. Thus, the $\SCond{\setx}{}{}$ modality binds
    the variables
    $\setx$ in $\phi$ and $\psi$,
just as $\forall x$ binds the occurrences of $x$ in $\forall x \phi$.
A typical formula in this language is
$\exists y (\SCond{x}{P(x,y)}{Q(x,y)})$, which can be read ``there is
some $y$ such that most $x$'s satisfying $P(x,y)$ also satisfy
$Q(x,y)$''.
 Note that we allow arbitrary nesting of
first-order and modal operators.
For simplicity, we assume that all variables used in formulas come from
the set $\{x_1, x_2, x_3, \ldots\}$.

To give semantics to $\LSCondfo(\Phi)$, we use {\em (first-order)
statistical plausibility structures} (over $\Phi$),
which generalize the semantics of statistical probabilistic
structures \cite{Bacchus,Hal4} and statistical preferential structures
\cite{Brafman97}.
Statistical plausibility structures over $\Phi$
are tuples of the
form $\PL = (\Dom, \pi, \Pl)$,
    where $\Dom$ is a domain, $\pi$ is an interpretation assigning each
    predicate symbol and function symbol in $\Phi$ a predicate or
    function of the right arity over $\Dom$, and
$\Pl$ is a plausibility measure on $\Dom^\infty$ (a countable product of
copies of $\Dom$) that satisfies one restriction, described below.
Note that
we can identify $\Dom^\infty$ with the set of all valuations by
associating a valuation $v$ with an infinite sequence $(d_1,
d_2, \ldots)$ of elements in $\Dom$, where $v(x_i) = d_i$.  Thus, we can
view $\Pl$
as defining a plausibility measure on the space of valuations.

We require that $\Pl$ treats all variables uniformly, in the following
sense:
\begin{itemize}
\item{REN.} If $h$ is a {\em finite permutation\/} of the natural
numbers (formally, $h: \IN \rightarrow \IN$ is a bijection such that
$h(n) = n$ for all but finitely many elements $n \in \IN$),
then $\Pl(A^h) = \Pl(A)$
for all $A \subseteq \Dom^\infty$, where $A^h = \{(d_{h(1)}, d_{h(2)},
d_{h(3)}, \ldots): (d_1, d_2, d_3, \ldots) \in A\}$.
\end{itemize}
REN assures us, for example, that if $A \subseteq \Dom$,
then $\Pl(A \times \Dom^\infty)$, the plausibility of a valuation giving
$x_1$ a value in $A$, is the same as the plausibility of a valuation
giving $x_2$ a value in $A$ (\ie $\Pl(\Dom \times A \times
\Dom^\infty)$) and, in fact, the
same as the plausibility of a valuation given $x_k$ a value in $A$, for
all $k$.  As the name suggests, REN guarantees that we can rename
variables, so that $\SCond{\setx}{\phi}{\psi}$ will be equivalent to
$\SCond{X[x/y]}{\phi[x/y]}{\psi[x/y]}$
if $y$ does not occur $\phi$ or $\psi$,
where $X[x/y]$ is the result
of replacing $x$ in $X$ by $y$ (if $x \in X$; otherwise $X[x/y] = X$).

\commentout{
\item{Prod1.} If $A, A' \subseteq Dom^n$ and $B \subseteq Dom^m$,
$\setx$ is a vector of $n$ variables, $\vec{y}$ is a vector of $m$
variables, and $x$ and $y$ are disjoint, and
$\Pl(\{v: v(\setx) \subseteq A\}) \le
\Pl(\{v: v(\setx) \subseteq A'\})$ then
\Pl(\{v: v(\setx,\vec{y}) \subseteq A \times B\}) \le
\Pl(\{v: v(\setx,\vec{y}) \subseteq A' \times B\})$.
\item{Prod2.} If $A, A' \subseteq Dom^n$ and $B \subseteq Dom^m$,
$\setx$ is a vector of $n$ variables, $\vec{y}$ is a vector of $m$
variables, and $x$ and $y$ are disjoint, then
$\Pl(\{v: v(\setx) \subseteq A\}) \le
\Pl(\{v: v(\setx) \subseteq A'\})$ iff
\Pl(\{v: v(\setx,\vec{y}) \subseteq A \times B\}) \le
\Pl(\{v: v(\setx,\vec{y}) \subseteq A' \times B\})$.
\end{itemize}
\end{description}
Prod1 and Prod2 say that, roughly
speaking, that variables can be treated independently.  If we take
$\Pl(A)$ for $A \subseteq \Dom^n$ to be an abbreviation of $\Pl(A \times
\Dom^\infty)$, then Prod1 says that $\Pl(A) \le \Pl(A')$ implies $\Pl(A
\times B) \le \Pl(A' \times B)$.  Prod2 strengthens this to an iff.
These constraints can be viewed as the generalizations of the
constraints imposed by Brafman \cite{Brafman97} (he called these
constraints permutation, weak concatenation, and strong concatenation,
respectively), who assumed the
plausibility was given in terms of a preferential order.%
\footnote{Actually, Brafman assumes that there is a separate
order defined on $\Dom^n$, for each finite $n$, rather than one order
defined on $\Dom^\infty$.  The two approaches are essentially
equivalent---we could have used either one here.  The connection to
valuations is perhaps clearer when we consider $\Dom^\infty$.}
These restrictions do not explicitly appear in the work on statistical
probabilistic reasoning \cite{Bacchus,Hal4}, since they follow from the
assumption that the probability measure on $\Dom^\infty$ is the product
measure derived from a probability on $\Dom$.

While these restrictions may all appear reasonable, note that Prod2 does
not apply to possibility measures, if we define the product possibility
measure in the most obvious way.  That is, if we start with a
possibility measure $\Poss$ on $\Dom$, define $\Poss^\infty$ on
$\Dom^\infty$ by taking $\Poss^\infty(d_1,d_2, \ldots) =
\inf_i\Poss(d_i)$, and taking $\Poss^\infty(A) = \sup_{\vec{d} \in
A}\Poss^\infty(\vec{d})$, then $\Poss^\infty$ satisfies Perm and Prod1,
but not Prod2.
}%
We may want to put a number of other restrictions on $\Pl$, to make it
act like a product measure, as Brafman \citeyear{Brafman97} does.  While
we believe such requirements may be quite reasonable, we do not make them
here, to simplify the presentation.  We discuss this issue further in
Section~\ref{statcase3}.

Given a statistical structure $\PL$ and a valuation $v$, we can
associate
with every formula $\phi$ a truth value in a straightforward way.
For an atomic formula such as $P(x,\boldc)$, we have
\begin{itemize}
\item $(\PL,v) \sat P(x,\boldc)$ if $(v(x),\pi(\boldc)) \in \pi(P)$.
\end{itemize}
Note that now we write $\pi(\boldc)$ rather than $\pi(w)(\boldc)$.  We
no longer have different worlds as we did in the subjective case.  Thus,
the issue of rigid vs.~nonrigid designators does not arise in the
statistical case.

We again treat quantification just as we do in first-order logic, so
\begin{itemize}
\item $(\PL,v) \sat \forall x \phi$ iff $(\PL,v') \sat \phi$ for
all $v' \sim_{x} v$.
\end{itemize}

The interesting case, of course, comes in giving semantics to formulas
of the form $\SCond{\setx}{\phi}{\psi}$.  In this case we have
\begin{itemize}\denselist
 \item $(\PL,v) \sat \SCond{\setx}{\phi}{\psi}$ if either
$\Pl(v': (\PL,v') \sat \phi, \, v' \sim_X v\}) =\bottom$
or
$\Pl(\{v': (\PL,v') \sat \phi \land
\psi, \, v' \sim_X v\}) >
\Pl(\{v': (\PL,v') \sat \phi \land \neg
\psi, \, v' \sim_X v\})$.
\end{itemize}
Again, we implicitly assume here that for each valuation $v$, vector
$\setx$ of variables, and formula $\phi$, the set of valuations
$\{v': (\PL,v') \sat \phi, \, v' \sim_X v\}$ is
in $\F$, the domain of $\Pl$.

As before, if $\phi$ is a sentence, then the truth of $\phi$ is
independent of $v$; thus, we write $\PL \sat \phi$ rather than $\PL,v
\sat \phi$.

\section{Axiomatizing first-order statistical conditional logic}
\label{statcase2}
We can axiomatize first-order statistical conditional logic in much the
same way as we did first-order subjective conditional logic.  Again, we
restrict attention to structures where the plausibility measures on
$\Dom^\infty$ are qualitative; let $\PQPLsfo$ be the class of all such
structures.

Let $\SysCsfo$ consists of all generalizations of the following axioms,
together with the inference rule MP.  In the axioms, we
write $\forall X \sigma$, where $X = \{x_1, \ldots, x_m\}$,
as an abbreviation for $\forall x_1 \ldots \forall x_m
\sigma$.
\begin{itemize}\denselist
 \item[C0$'$.] All instances of valid formulas of first-order logic with
equality
 \item[C1$'$.] $\SCond{\setx}{\phi}{\phi}$
 \item[C2$'$.]
$((\SCond{\setx}{\phi}{\psi_1})\land(\SCond{\setx}{\phi}{\psi_2}))
\rimp   (\SCond{\setx}{\phi}{\psi_1\land\psi_2})$
 \item[C3$'$.]
$((\SCond{\setx}{\phi_1}{\psi})\land(\SCond{\setx}{\phi_2}{\psi}))
\rimp
    (\SCond{\setx}{(\phi_1\lor\phi_2)}{\psi})$
 \item[C4$'$.]
$(\SCond{\setx}{\phi_1}{\phi_2}\land \SCond{\setx}{\phi_1}{\psi})
\rimp
    \SCond{\setx}{\phi_1\land\phi_2} {\psi}$
 \item[R1$'$.] $\forall \setx(\phi_1 \dimp \phi_2) \rimp
((\SCond{\setx}{\phi_1}{\psi} \rimp (\SCond{\setx}{\phi_2}{\psi})$
\item[R2$'$.] $\forall \setx(\psi_1 \rimp \psi_2) \rimp
((\SCond{\setx}{\phi}{\psi_2} \rimp (\SCond{\setx}{\phi}{\psi_2})$
\item[U.] $\forall X \psi \rimp (\SCond{\setx}{\phi}{\psi})$
\item[Ren.] $\SCond{\setx}{\phi}{\psi} \rimp
\SCond{X[x/y]}{\phi[x/y]}{\psi[x/y]}$, if $y$ does not occur in $\phi$ or
$\psi$.
\end{itemize}

As the notation suggests, C1$'$--C4$'$, R1$'$, and R2$'$ are the obvious
analogues C1--C4, R1, and R2, except that R1$'$ and R2$'$ are now axioms
rather than inference rules.  C0$'$ subsumes C0 and F1--F5 in $\SysCfo$.
Note that we no longer need a special notion of substitutivity; there is
only one world, and there are no concerns regarding the substitution of
nonrigid terms into a modal context.  For similar reasons, there is no
analogue of F6 and F7 here.  Ren and U are analogues of similar axioms
for statistical probabilistic structures \cite{Bacchus,Hal4}; here we
need to require REN to ensure that Ren holds.

\thm\label{statcompleteness}
$\SysCsfo$ is a sound and complete axiomatization of $\LCondsfo$
with respect to
$\PQPLsfo$. \ethm
\prf The basic idea similar to that of the proof of
Theorem~\ref{subjcompleteness}; indeed, the proof is even simpler.
See the appendix for details. \eprf

\section{Other approaches to first-order statistical
conditional logic}\label{statcase3}
We have already remarked that we can construct ``statistical''
first-order analogues of all the approaches considered in the
propositional case.  We omit the formal definitions here.  Let
$\Pwprefsfo,\Pprefsfo,\Pkappasfo,\Pposssfo,$ and $\Pepssfo$ be
    the subsets of $\PQPLsfo$ that correspond to
    well-founded preference orderings,
    preference orderings, $\kappa$-rankings, possibility measures,
    and PPDs, respectively.
The results are similar to those in Section~\ref{other.approaches},
so we just sketch them here.

With $\kappa$-rankings and possibility
measures, we need to require the obvious analogues of C6 and C7, namely
\begin{itemize}
\item[C6$'$.]
$(\SCond{\setx}{\phi}{\psi}) \land \neg(\SCond{\setx}{\phi}{\neg\xi})
\rimp \SCond{\setx}{\phi \land \xi}{\psi}$
\item[C7$'$] $\neg(\SCond{\setx}{\true}{\false})$
\end{itemize}
As we would expect, $\epsilon$-semantics satisfies C7$'$ but not
necessarily C6$'$.%
\footnote{Brafman \citeyear{Brafman97} discusses {\em pointed\/} PPDs,
in which all the relevant limits are guaranteed to exist; for pointed
PPDs, C6$'$ holds as well.}

We have the following analogue of Theorem~\ref{sound+complete}.
\thm\label{eps2} $\SysCsfo$+{\rm C7$'$} is a sound and complete
axiomatization of
$\LCondsfo$ with respect to $\Pepssfo$. \ethm
\prf Follows from the proof of Theorem~\ref{statcompleteness} using the
same techniques as those used to prove Theorem~\ref{sound+complete} from
Theorem~\ref{subjcompleteness}.  We omit further details here.
\eprf

Statistical plausibility structures based on well-founded preferential
structures and $\kappa$-rankings also satisfy the following analogue of
$\forall 3$:
\begin{itemize}
\item[$\forall 3'$.] $\forall y (\SCond{\setx}{\phi}{\psi}) \rimp
(\SCond{\setx}{\phi}{\forall y \psi})$
if $y$ does not occur free in $\phi$ or in $\setx$.
\end{itemize}

Interestingly, Brafman~\citeyear{Brafman97} shows that $\SysCsfo$
together with C6$'$, C7$'$, and $\forall 3'$ is complete with respect to
totally-ordered well-founded preferential structures.%
\footnote{Actually, there are a number of minor differences between the
framework we have presented and that of Brafman. For example, Brafman
assumes that there is a separate
order defined on $\Dom^n$, for each finite $n$, rather than one order
defined on $\Dom^\infty$.  The two approaches are essentially
equivalent---we could have used either one here.  The connection to
valuations is perhaps clearer when we consider $\Dom^\infty$.  He also
has the axiom $(\SCond{\setx}{\phi}{\psi}) \rimp (\exists x \phi \rimp
\exists x (\phi \land \psi))$ instead of C7$'$.  It is not hard to show
that these axioms are equivalent in the presence of all the other
axioms.}
These are essentially identical to the structures generated by
$\kappa$-rankings.  Thus, we have the following result.
\thm\label{Brafthm} {\rm \cite{Brafman97}}
$\SysCsfo+\{${\rm C6$'$, C7$'$, $\forall 3'$}$\}$ is a sound and
complete
axiomatization of $\LCondsfo$ with respect to $\Pkappasfo$. \ethm

In light of Brafman's result, it seems likely that
$\SysCsfo$+$\{${\rm C6$'$, $\forall 3'$}$\}$ is a sound and complete
axiomatization of $\LCondsfo$ with respect to $\Pwprefsfo$, although we
have not checked details.

Just as in the subjective case, $\forall 3'$ is not valid in
statistical possibility structures or
(non-well-founded) preferential structures, but a variant of the crooked
lottery example does give us a valid formula for these structures too
that does not follow from $\SysCsfo$+ $\{${\rm C6$'$, C7$'$}$\}$.

Up to now, we have put minimal structure on the plausibility measure on
$\Dom^\infty$.  In the case of statistical probability structures, the
probability measure was assumed to be the product measure induced by a
probability measure on $\Dom$.  We can make an analogous assumption in
the case of $\epsilon$-semantics, possibility measures, and
$\kappa$-rankings.  For example, if we start with a
possibility measure $\Poss$ on $\Dom$, we can define $\Poss^\infty$ on
$\Dom^\infty$ by taking $\Poss^\infty(d_1,d_2, \ldots) =
\inf_i\Poss(d_i)$, and taking $\Poss^\infty(A) = \sup_{\vec{d} \in
A}\Poss^\infty(\vec{d})$ for $A \subseteq \Dom^\infty$.  A similar
construction works for $\kappa$-rankings, except $\inf$ is replaced by
$+$ and $\sup$ is replaced by $\min$.  We get extra properties if we
assume such a product measure construction, although the exact
properties depend on the underlying notion of likelihood that we start
with.  For example, one property we get in all cases is the following:
\begin{itemize}
\item If $A, A' \subseteq Dom^n$ and $B \subseteq Dom^m$,
$\vec{y} = \<y_1, \ldots, y_n\>$, $\vec{z} = \<z_1, \ldots, z_m\>$,
$\vec{y}$ and $\vec{z}$ are disjoint,
$\Pl(\{v: (v(y_1), \ldots, v(y_n)) \in A\}) \le
\Pl(\{v: (v(y_1), \ldots, v(y_n)) \in A'\})$ then
$\Pl(\{v: (v(y_1), \ldots, v(y_n), v(z_1), \ldots, v(z_m)) \in A \times B\})
\le
\Pl(\{v: (v(y_1), \ldots, v(y_n), v(z_1), \ldots, v(z_m)) \in A' \times
B\})$.
\end{itemize}
This property is captured by the axiom
$$\SCond{\setx}{\phi}{\psi} \rimp \SCond{\setx}{\phi \land
\phi'}{\psi},$$
where the set of variables free in $\phi'$ is disjoint from the set of
variables free in $\phi \land \psi$.

Whether or not we assume that $\Pl$ is generated as a product measure
somehow, once we have $\forall 3'$ as an axiom (or the closely related
variant as in the crooked lottery example), we get the problems in
the statistical case similar to those we saw in the subjective case.
For example, suppose $\forall 3'$ is valid.
Consider the statement
$$\forall y (\SCond{x}{\True}{\neg\Married(x,y)}) \label{married1}.$$
This states that for any individual $y$, most individuals are not
    married to $y$. This seems reasonable since each $y$ is married to
    at most one individual, which clearly constitutes a small fraction of the
    population.
$\forall 3'$ then gives us
$$
\SCond{x}{\True}{\forall y \neg\Married(x,y)}.
$$
That is, most people are not married! This certainly does not seem to be
a reasonable conclusion.

It is straightforward to construct similar
examples for the statistical variants of the other approaches, again,
with the exception of plausibility structures and
    $\epsilon$-semantics.
We note that these problems occur for precisely the same reasons they
    occur in the subjective case. In particular,
$\forall 3'$ holds whenever the plausibility measure on $\Dom^\infty$
satisfies A2$^*$.

This shows that, just as for the subjective case, we need
the greater generality of plausibility measures and $\epsilon$-semantics
to correctly model first-order statistical reasoning about conditionals.

We observe that problems similar to the lottery paradox occur in the
    approach of Lehmann and Magidor \citeyear{LehmannMagidor90}, which
    can be viewed as a hybrid of subjective and statistical
    conditionals based on on preferential structures.
More precisely, rather than putting a preferential ordering on worlds or
on valuations, they put an ordering on world-valuation pairs.  While
this greater flexibility allows them to avoid some problems associated
with putting an order solely on worlds or on valuations, the fundamental
difficulty still remains.

 Finally, we observe that the approach of \cite{Schlechta95}, which
is
    based on a novel representation of ``large'' subsets, is
in the spirit of our notion of statistical defaults (although his
language is somewhat less expressive than ours).

\section{Discussion}\label{discussion}
We have
considered a number of different approaches to ascribing
semantics to both a subjective and statistical
first-order logic of conditionals in a number of ways.
Our analysis
shows that,
once we move to the first-order case, significant differences arise
between approaches that were shown to be equivalent in the propositional
case.
This vindicates the intuition that there are significant differences
between these approaches, which the propositional language is simply
too weak to capture.
The analysis also supports our choice of plausibility structures as the
semantics for first-order conditional logic; it shows that, with the
exception of $\epsilon$-semantics, all the previous approaches
have significant shortcomings, which manifest themselves in
lottery-paradox type situations.
Plausibility also lets us home in on what properties of an
approach give us lead to an infinitary AND rule like $\forall 3$.%
\commentout{
\footnote{As another example, we can show that A2$^*$ together with
A3$^*$, an infinitary version of A3 (namely, if $\Pl(A_i) = \bottom$ for
all $i \in I$, then $\Pl(\union_{i \in I} A_i) = \bottom$) gives us an
infinitary analogue of the OR rule:
$$\forall x (\phi \Cond \psi) \rimp ((\exists x \phi) \rimp \psi),$$ if
$x$ does not appear free in $\psi$.}
}

What does all this say about default reasoning?
As we have argued, statements like ``birds typically fly'' should perhaps be
thought of as statistical statements, and should thus be represented as
$\SCond{x}{\Bird(x)}{\Fly(x)}$.  Such a representation gives us a logic
of defaults, in which statements such as ``birds typically fly'' and
``birds typically do not fly'' are inconsistent, as we would expect.

Of course, what we really want to do with such typicality statements is to
draw default conclusions about individuals.
Suppose we believe such a typicality
statement.  What other beliefs should follow?
In general,
$\forall x (\Bird(x) \Cond \Fly(x))$ does not follow; we should not
necessarily believe that {\em all\/} birds are likely to fly.  We may
well know that Tacky the penguin \cite{Tacky} does not fly .  As long as
Tacky is a rigid designator, this is simply inconsistent with believing
that all birds are likely to fly.  In the absence of information about
any particular bird, $\forall x (\Bird(x) \Cond \Fly(x))$ may well be
a reasonable belief to hold.  Moreover, no matter what we know about
exceptional birds, it seems reasonable to believe
$\SCond{x}{\True}{(\Bird(x) \Cond \Fly(x))}$: almost all birds are likely
to fly (assuming we have a logic that allows the obvious combination of
statistical and subjective plausibility).

Unfortunately, we do not have a general approach that will let us go
{f}rom believing that birds typically fly to believing that almost all
birds are likely to fly.  Nor do we have an approach that allows us to
conclude that Tweety is likely to fly given that birds typically fly and
Tweety is a bird (and that we know nothing else about Tweety).
These issues were
addressed in the first-order setting by both Lehmann and Magidor
\citeyear{LehmannMagidor90}
and Delgrande \citeyear{Delgrande}.  The key feature
of their approaches, as well as other propositional approaches
rests upon getting a suitable notion of irrelevance.
While we also do not have a general solution to the problem of
irrelevance, we believe
that plausibility structures give us the tools to
study it in an abstract setting.  We suspect that many of the intuitions
behind probabilistic approaches that allow us to cope with irrelevance
\cite{BGHKfull,KH.irrel} can also be brought to bear here.
We hope to return to this issue in future work.

\appendix
\section{Proofs}

\othm{subjcompleteness}
$\SysCfo$ is a sound and complete axiomatization of $\LCondfo$
with respect to
$\PQPLfo$. \eothm
\prf A formula $\phi$ is said to be {\em consistent\/} with $\SysCfo$ if
$\SysCfo \not\vdash \neg \phi$.  A finite set of formulas $\{\sigma_1,
\ldots, \sigma_k\}$ is consistent with $\SysCfo$ if their conjunction
$\sigma_1 \land \ldots \land \sigma_k$ is consistent with $\SysCfo$.  An
infinite set $\Sigma$ of formulas is consistent with $\SysCfo$ if every
finite subset of $\Sigma$ is consistent with $\SysCfo$.  Finally, a set
$\Sigma$ is said to be a {\em maximal consistent set of sentences\/} if
(1) it consists only of sentences (recall that a sentence is a formula
with no free variables), (2) it is consistent and (3) no strict superset
of $\Sigma$ consisting only of sentences is consistent.
In the discussion below, all maximal consistent sets are maximal
consistent sets of sentences; however, the other consistent sets we
construct may include formulas that are not sentences.

Our goal is to show that a formula $\phi$ is consistent with
$\SysCfo$ iff it is satisfiable in a first-order plausibility structure.
As usual, this clearly suffices to prove completeness.  We can also
assume without loss of generality that $\phi$ is a sentence, for using
standard arguments of first order logic (see \cite[p.~109]{Enderton}) we
can show that if $y_1, \ldots, y_m$ are the free variables in $\phi$,
then $\phi$ is provable iff its universal closure $\forall y_1 \ldots
\forall y_m \phi$ is provable, and hence $\phi$ is consistent iff
$\exists y_1  \ldots \exists y_m \phi$ is consistent.

Let $\C$ be a countable set
of constant symbols not in $\Phi$, let
$\Phi'$ consist of the
symbols in $\Phi$ that actually appear in $\phi$, and let
$\Phi^+ = \Phi' \union \C$.%
\footnote{This proof would go through without change if we took $\Phi^+$
to be $\Phi \union \C$.  However, for the proof of
Theorem~\ref{sound+complete}, it is useful to restriction attention to a
language that is guaranteed to be countable.}
As usual in Henkin-style completeness proofs, we construct a
structure satisfying $\phi$ using maximal consistent subsets of
$\SysCfo$ (in the language $\LCondfo(\Phi^+)$).

A maximal consistent subset $A$ of $\SysCfo$ is said to be {\em
$\C$-good\/} if
(1) $\neg \forall x \psi \in  A$ implies $\neg \psi[x/\boldc] \in A$
for some $\boldc \in \C$ and (2) $\forall x \psi \in A$ implies
$\psi[x/\boldc] \in A$ for all $\boldc \in \C$.  Note that property (2)
holds automatically for maximal consistent sets in first-order logic,
but does not hold in general in our logic, because of our restriction on
F1.  Intuitively, $A$ is
$\C$-good if the constants in $\C$ are rigid designators such that every
domain element is the interpretation of some constant in $\C$.

The proof now proceeds according to the following steps:
\begin{enumerate}
\item We show that there is a $\C$-good maximal consistent set $C^*$ that
includes
$\phi$. This follows closely the standard Henkin-style completeness
proof for first-order logic \cite{Enderton}.
\item We construct a structure $\PL$ by using the formulas in $C^*$.
This
step uses techniques from \cite{Enderton} for defining the domain, and
from \cite{FrH5Full} for defining the set of possible worlds and
the plausibility measure over them
\item We show that $\Pl \sat \phi$.
Again, this argument is in the spirit of the standard
Henkin-style completeness for first-order logic.
\end{enumerate}
For the first step,
we proceed as follows.
Let $\sigma_0, \sigma_1,\ldots$ be an enumeration of the formulas in
the language $\LCondfo(\Phi^+)$.
We inductively construct a sequence $A_0, A_1,
\ldots$
of finite sets of formulas such that $A_n$ is consistent.  Let $A_0 =
\{\phi\}$.   Let $A_{k+1}$ consist of $A_k$ together with the formula
$\neg \forall x \sigma_{k+1} \rimp \neg \sigma_{k+1}[x/\boldc]$, where
$\boldc$ is a constant in $\C$ that does not appear in any of the
formulas in
$A_k$ or in $\sigma_{k+1}$.  (This is possible since $A_k$ is finite.)
Intuitively,
$\neg \forall x \sigma_{k+1} \rimp \neg \sigma_{k+1}[x/\boldc]$ says
that
$\boldc$ provides a witness to the fact that $\neg \sigma_{k+1}$ does
not hold for all $x$, if such a witness is necessary.

We claim that $A^* = \union_k A_k$
is consistent.
This follows from the following somewhat
more general lemma.
\lem\label{witness}  If $B_0$ is a finite consistent set of formulas,
and for $k \ge 0$, $B_{k+1} = B_k \union \{\neg \forall x \sigma \rimp
\sigma[x/\boldc]\}$ for some formula $\sigma$ and constant $\boldc$ that
does not appear in $B_k$ or $\sigma$, then $\union_k B_k$ is consistent.
\elem
\prf From the definition of consistency, it clearly suffices to prove
that $B_k$ is consistent for all $k \ge 0$.  We do this by induction on
$k$. By assumption
$B_0$ is consistent. Suppose
$B_k$ is consistent but $B_{k+1}$ is not.  Suppose $B_{k+1} = B_k \union
\{\neg \forall x \sigma \rimp \neg \sigma[x/\boldc]\}$.   Identifying
$B_k$ with the conjunction of formulas in $B_k$, it then follows that
$$\SysCfo \vdash B_k \rimp (\neg \forall x \sigma \land
\sigma[x/\boldc]).$$
Since $\boldc$ does not appear in any of the formulas in $B_k$ or
$\sigma$, a standard argument from first-order logic (see
\cite[p.~116]{Enderton}) can be used to show that
$$\SysCfo \vdash B_k \rimp (\neg \forall x \sigma \land
\forall x \sigma),$$
contradicting the consistency of $B_k$. \eprf

Let $B^*$ consist of all the formulas in $A^*$ together with all the
formulas of the form $\forall x \sigma \rimp \sigma[x/\boldc]$, where
$\sigma$ is a formula in $\LCondfo$ and $\boldc \in \C$.  We claim
that $B^*$ is consistent.  For suppose not.  Then there exists a finite
set of formulas in $A^*$, say $A'$, and a finite set of formulas $B'
\subseteq \LCondfo$, and a finite set of constants $\C' \subseteq \C$
such that
$$\SysCfo \vdash A' \rimp ((\land_{\sigma \in B'} \forall x \sigma)
\land
(\lor_{\sigma \in B',\boldc \in \C'} \neg \sigma[x/\boldc])).$$
Suppose $\C' = \{\boldc_1, \ldots, \boldc_k\}$.  Let $y_1, \ldots, y_k$
be fresh variables, that do not appear in the formulas in $A'$ or $B'$.
Let $A''$ and $B''$ be the result of replacing all occurrences of
$\boldc_1, \ldots, \boldc_k$ in the formulas in $A'$ and $B'$,
respectively, by the variables $y_1, \ldots, y_k$.  Again, using
standard techniques \cite[p.~116]{Enderton}, we have
$$\SysCfo \vdash A'' \rimp ((\land_{\sigma \in B''} \forall x \sigma)
\land
(\lor_{\sigma \in B'', y \in \{y_1, \ldots, y_k\}} \neg \sigma[x/y])).$$
It follows from F1 that
$(\land_{\sigma \in B''} \forall x \sigma) \land
(\lor_{\sigma \in B'', y \in \{y_1, \ldots, y_k\}} \neg \sigma[x/y])$ is
inconsistent.  Thus, $A''$ must be inconsistent.  But it follows from
Lemma~\ref{witness} that $A''$ is consistent.  This gives us the desired
contradiction.

Let $B^{\dagger}$ consist of all the sentences in $B^*$.
Using standard techniques, we can extend
$B^{\dagger}$ to a maximal
consistent
set of sentences:
We construct a sequence $C_0, C_1,
\ldots$ of consistent sets of sentences by taking $C_0 = B^{\dagger}$
and $C_{k+1}
= C_k \union \{\sigma_k\}$ if $C_k \union \{\sigma_k\}$ is consistent
and $\sigma_k$ is a sentence, and $C_{k+1} = C_k$ otherwise.  Let $C^* =
\union_{k=1}^\infty C_k$.  $C^*$ is then easily seen to be
a maximal consistent set of sentences.
Moreover, our construction guarantees that
$C^*$ is $\C$-good and contains $\phi$. This completes Step 1 of the
proof.

We now proceed to the second step of the proof, where we construct a
first-order subjective plausibility structure based on $C^*$. First, however,
we need two more definitions in order to allow us to characterize the
domain and the set of possible worlds in our desired
plausibility
structure.
\begin{itemize}
\item We define an equivalence relation $\sim$ on $\C$ by defining
$\boldc
\sim \boldc'$ if $\boldc = \boldc' \in C^*$.  Let $[\boldc] = \{\boldc':
\boldc \sim \boldc\}$.  As we shall see, these equivalence classes will
be the domain elements in our structure.
\item If $A$ is a set of formulas, define
$A/\PBox = \{\psi: \PBox \psi \in A\}$.  The worlds in our
structure will be all $\C$-good maximal consistent sets $A$ of
sentences such that $A/\PBox = C^*/\PBox$.
\end{itemize}

We want to ensure that global properties, such as equality of domain
elements and conditional statements, are true in all worlds in the
structure. As we shall see, our construction is such that
formulas in $C^*/\PBox$ are true in all these worlds. Thus, we need the
following lemma.

\lem\label{needit} If $\psi$ is of the form $\psi' \Cond \psi''$ or
$\boldc = \boldc'$, for $\boldc, \boldc' \in \C$
and $A$ is a $\C$-good maximal consistent set,
$\psi \in A$ iff $\psi \in A/\PBox$. \elem
\prf
Suppose $A$ is a $\C$-good maximal consistent set of formulas and
$\psi$ is of the form $\psi' \Cond \psi''$.  If $\psi \in
A$, then $\PBox \psi \in A$ by C5, so $\psi \in A/\PBox$.
Conversely, if $\psi \in A/\PBox$, then $\PBox \psi \in A$.
Suppose, by way of contradiction, that $\psi \notin A$.  Since $A$
is a maximal consistent set, we must have $\neg \psi \in A$.  By C5,
it follows that $\PBox \neg \psi \in A$.  That is, both $\psi \Cond
\false$ and $\neg \psi \Cond \false$ are in $A$.  By C3, it follows
that $\true \Cond \false \in A$.  By RW, we have that both $\true
\Cond \psi'$ and $\true \Cond \psi''$ are in $A$.  By C4, we have that
$\psi = \psi' \Cond \psi'' \in A$, contradicting our assumption.

Now suppose $\psi$ is of the form $\boldc = \boldc'$.  By F6, we have
that $\forall x, y (x=y \rimp \PBox(x=y)) \in A$.  Since $A$ is
$\C$-good, it follows that $\boldc = \boldc' \rimp \PBox(\boldc =
\boldc') \in A$.  Similarly, by F7, we have that $\boldc \ne \boldc'
\rimp \PBox(\boldc \ne \boldc') \in A$.  We can now show that $\psi
\in A$ iff $\psi \in A/\PBox$ just as we did in the case of $\psi'
\Cond \psi''$, replacing the use of C5 by these consequences of F6 and
F7. \eprf

We construct a first-order subjective plausibility structure
$\PL = (\Dom,W, \Pl, \pi)$ as follows:
\begin{itemize}
\item $\Dom = \{[\boldc]: \boldc \in \C\}$
\item $W =\{w :$ $w$ is a $\C$-good maximal consistent set of
sentences and
$w/\PBox = C^*\/PBox$ (\ie $\PBox \psi \in w$ iff $\PBox \psi \in
C^*$) $\}$

\item $\pi$ is defined so that $\pi(w)(\boldc) = [\boldc]$ for the
constants $\boldc \in \C$ and $\pi(w)$ for the
symbols in $\Phi'$ is determined by the atomic sentences in $w$ in the
obvious way
(see \cite[p.~131]{Enderton}).  For example, if $P$ is a binary
predicate, then $([\boldc],[\boldc']) \in \pi(w)(P)$ iff
$P(\boldc,\boldc')$ is one of the formulas in $w$.
\item $\Pl$ is defined so that for all formulas
$\psi$ and $\psi'$, we have $\Pl([\psi]) \le \Pl([\psi'])$ iff
$(\psi \lor \psi') \Cond \psi' \in C^*$, where $[\sigma] = \{w:
\sigma \in w\}$.
\end{itemize}
In their completeness proof for propositional
conditional logic, Friedman and
Halpern \citeyear{FrH5Full} define a plausibility measure $\Pl$ in a
similar way and show that it is well defined (\ie if $[\psi] = [\psi']$
and $[\sigma] = [\sigma']$, then $(\psi \lor \sigma) \Cond \sigma \in
C^*$ iff $(\psi' \lor \sigma') \Cond \sigma' \in C^*$) and that it
satisfies A1, A2, and A3, so that $\Pl$ is a qualitative plausibility
measure.  (See the proof of Theorem 8.2 in \cite{FrH5Full}.)  The same proof
applies here without change, so we do not repeat it.

Finally, we move to the third and last step of the proof: showing that
$\PL$ satisfy $\phi$. To do so we
prove the standard {\em truth lemma}, namely
\begin{quote}
{\rm ($*$)} $(\PL,w) \sat \psi \mbox{ iff } \psi \in w$,
for all $w \in W$.
\end{quote}
We prove (*) by a straightforward induction on the depth of nesting of
$\Cond$ in $\psi$, with a subinduction on structure.

If $\psi$
is an atomic formula of the form $\boldc =\boldc'$, this follows from
Lemma~\ref{needit} and the definition of $\pi$.  For other atomic
formulas, this is immediate from the definition of $\pi$.

If $\psi$ is a conjunction or a negation, it is immediate from the
induction hypothesis.

If $\psi$ has the form $\psi' \Cond \psi''$,
then we have
$$\begin{array}{llr}
&(\PL,w) \sat \psi' \Cond \psi''\\
\mbox{iff } &\Pl(\intension{\psi' \land \psi''}) > \Pl(\intension{\psi'
\land \neg \psi''}) \mbox{ or } \Pl(\intension{\psi'}) = \bot\\
\mbox{iff } &\Pl([\psi' \land \psi'']) > \Pl([\psi'
\land \neg \psi'']) \mbox{ or } \Pl([\psi']) = \bot &\mbox{[induction
hypothesis]}\\
\mbox{iff } &((\psi' \land \psi'') \lor (\psi' \land \neg \psi'')) \Cond
(\psi' \land \psi'') \in C^* \mbox{ or } \psi' \Cond \false \in C^*
&\mbox{[by definition of $\Pl$]}\\
\mbox{iff } &\psi' \Cond \psi'' \in C^* \mbox{ or } \psi' \Cond \false
\in C^* &\mbox{[by LLE,C1,C2]}\\
\mbox{iff } &\psi' \Cond \psi'' \in C^* &\mbox{[by RW]}\\
\mbox{iff } &\PBox(\psi' \Cond \psi'') \in C^* &\mbox{[by Lemma~\ref{needit}]}\\
\mbox{iff } &\PBox(\psi' \Cond \psi'') \in w &\mbox{[since $w \in W$]}\\
\mbox{iff } &\psi' \Cond \psi'' \in w &\mbox{[by Lemma~\ref{needit}]}
\end{array}$$

Finally, if $\psi$ has the form $\forall x \psi'$, then we have
$$\begin{array}{llr}
&(\PL,w) \sat \forall x \psi\\
\mbox{iff } &(\PL,w) \sat \psi[x/\boldc] \mbox{ for all $\boldc \in
\C$}\\
\mbox{iff } &\psi[x/\boldc] \in w \mbox{ for all $\boldc \in
\C$} &\mbox{[induction hypothesis]}\\
\mbox{iff } &\forall x \psi \in w &\mbox{[since $w$ is $\C$-good]}
\end{array}
$$

This completes the proof of ($*$).  It now follows that $(\PL,C^*) \sat
\phi$.  We can easily get from this a structure over the vocabulary
$\Phi$ that satisfies the formula $\phi$: we simply define $\pi$ in an
arbitrary way for the symbols in $\Phi - \Phi'$, and ignore the
interpretation of the symbols in $\C$.  This completes the proof of
the theorem.
\eprf

\bigskip

\othm{sound+complete}
$\SysCfo+${\rm C7} is a sound and complete axiomatization
of $\LCondfo$
with respect to $\Pepsfo$.
\eothm

\medskip

\prf As in the proof of Theorem~\ref{subjcompleteness},
it suffices to
show that if $\phi$ is consistent with $\SysCfo+${\rm C7}, then it is
satisfiable in  a structure in $\Pepsfo$.  The first steps in the proof
mimic those of the proof of Theorem~\ref{subjcompleteness}.  We
define $\Phi^+ = \Phi' \union \C$ as in the proof of
Theorem~\ref{subjcompleteness}.
We then construct a plausibility structure $\PL = (D,W,\Pl,\pi)$
satisfying $\phi$
by considering maximal ($\SysCfo+${\rm C7})--consistent sets of
sentences.
It is easy to see that the structure $\PL$ is what is called in
\cite{FrH5Full} (following \cite{Lewis73}) {\em normal\/}: we must have
$\Pl(W) > \bot$ (otherwise C7 would not be valid in $\PL$).  By
Theorem~6.3 in \cite{FrH5Full}, it follows that there is a PPD $PP$ on $W$
that satisfies the same defaults.  More precisely, if $\PL_{pp} =
(D,W,\Pl_{PP},\pi)$,
where $\Pl_{PP}$ is the plausibility measure
corresponding to $PP$, as described in Section~\ref{prop-logic}, then we
have
$(\PL,w) \sat \psi \Cond \psi'$ iff $(\PL_{PP},w) \sat \psi \Cond
\psi'$.%
\footnote{Theorem~6.3 in \cite{FrH5Full} applies only to countable
languages, so it is important that we use $\Phi^+$ here, rather than
$\Phi \union \C$, which may not be countable.}
A straightforward induction on the structure of formulas now shows that
$\PL$ and $\PL_{PP}$ agree on all sentences in $\LCondfo(\Phi^+)$.  This
gives us the desired PPD structure satisfying $\phi$, and completes the
proof. \eprf

\bigskip

\opro{A2*}
A2$^*$ holds in every plausibility structure in $\Pwpreffo$ and
$\Pkappafo$.
\eopro

\medskip

\prf We start with $\Pkappafo$.
Suppose $\PL = (D,W,\Pl_{\kappa},\pi) \in \Pkappafo$, where
$\kappa$ is the ranking to which $\Pl_{\kappa}$ corresponds.  Since
lower ranks correspond to greater plausibility, we have $\kappa(A) \le
\kappa(B)$ iff $\Pl_\kappa(A) \ge \Pl_\kappa(B)$.  Let
$\{A_i: i
\in I\}$ be a collection of pairwise disjoint sets such that $\kappa(A -
A_i) < \kappa(A_i)$ for all $i \in I - \{0\}$.  We claim that
(1) $\kappa(A)
= \kappa(A_0)$, (2) $\kappa(A) < \kappa(A_i)$ for $i \in I - \{0\}$,
and (3) $\kappa(A_0) < \kappa(\union_{i \in I - \{0\}} A_i)$.
(2) follows immediately from the assumption that $\kappa(A - A_i) <
\kappa(A_i)$, since $\kappa(A) \le \kappa(A - A_i)$.  (1)
follows from (2) and the observations that (a) $\kappa(A) = \min_{i \in
I} \kappa(A_i)$ and (b) the range of $\kappa$ is the natural numbers.
Finally, (3) follows from (1) and (2) and the observation that
$\kappa(\union_{i \in I - \{0\}} A_i) = \min_{i \in I - \{0\}}
\kappa(A_i)$.  From (1), (2), and (3), it is immediate that $\kappa(A_0)
< \kappa(A - A_0)$.

The argument in the case of $\Pwpreffo$ is similar in spirit.
Suppose $\PL = (D,W,\Pl_{\prec},\pi) \in \Pwpreffo$, where $\Pl_{\prec}$
is constructed from a partial order $\prec$ on $W$ as described
in Section~\ref{prop-logic}.  Again, let
$\{A_i: i
\in I\}$ be a collection of pairwise disjoint sets such that
$\Pl_{\prec}(A
- A_i) < \Pl_{\prec}(A_i)$ for all $i \in I - \{0\}$.
Recall that
$\Pl_\prec(A) \le \Pl_\prec(B)$ if and only if for all $w \in A - B$,
    there is a world $w' \in B$ such that $w' \prec w$ and there is no
    $w'' \in A - B$ such that $w'' \prec w'$.
Thus, to show that $\Pl_\prec(A_0) \ge \Pl_\prec(A-A_0)$, we must show
that if $w \in A
- A_0$, then there exists some $w' \in A_0$ such that $w'
\prec w$ and there is no $w'' \in A - A_0$ such that $w'' \prec w'$.
Suppose not.  Then we construct an infinite decreasing sequence $\ldots
w_k \prec w_{k-1} \prec \cdots \prec w_0$, contradicting the assumption
that $\prec$ is well founded.  We proceed as follows.  Let $w_0 = w$.
Suppose inductively we have constructed $w_0, \ldots, w_k$.  If $w_k \in
A_0$, by assumption, there is some $w_{k+1} \in A-A_0$ such that
$w_{k+1} \prec w_k$.  If $w_k \in A-A_0$, then
$w_k \in A_i$ for some $i \ne 0$.  Since $\Pl_\prec(A-A_i) >
\Pl_\prec(A_i)$, it follows from the construction of $\Pl_\prec$ that
there
must be some $w_{k+1} \in A-A_i$ such that $w_{k+1} \prec w_k$.  This
completes the inductive proof, and gives us the desired contradiction.
It is easy to see that because $A_0$ and $A-A_0$ are disjoint, the fact
that $\Pl_\prec(A_0) \ge \Pl_\prec(A-A_0)$ implies that
$\Pl_\prec(A_0) > \Pl_\prec(A-A_0)$, as desired.
\eprf

\bigskip

\opro{forall3} $\forall3$ is valid in all plausibility structures
satisfying A2$^*$. \eopro

\medskip

\prf  Suppose $\PL = (D,W,\Pl,\pi)$ satisfies A2$^*$ and
$(\PL,w,v)  \sat \forall x (\phi \Cond \psi)$, where $x$ does not appear
free in $\phi$.  It follows that
\begin{equation}\label{eqpl1}
\intension{\phi \land \psi}_{\PL,w,v'} >
\intension{\phi \land \neg \psi}_{\PL,w,v'}, \mbox{ for all valuations
$v'$ such that $v' \sim_x v$}.
\end{equation}
Let $A = \intension{\phi}_{\PL,w,v}$.  (Note that we have $A =
\intension{\phi}_{\PL,w,v'}$ for all $v'$ such that $v' \sim_x v$,
since $x$ is not free in $\phi$.)
For each $d \in D$, let $A_d = \intension{\phi \land \neg
\psi}_{\PL,w,v_d}$, where $v_d \sim_x v$ and
$v_d(x) = d$.  Let $A' =
\intension{\phi \land \forall x \psi}$.  Note that $A = A' \union
(\union_{d \in D} A_d)$ and that $A'$ is disjoint from $A_d$, for each
$d
\in D$.  The sets $A_d$ are not necessarily disjoint.  Thus, let $B_d$
be such that $B_d \subseteq A_d$, the sets $B_d$ are pairwise disjoint,
and $\union_{d \in D} B_d = \union_{d \in D} A_d$.  (We can always find
such sets $B_d$.  If $D$ is countable, say $D = \{1, 2, 3, \ldots\}$
without loss of generality, then we can take $B_1 = A_1$ and $B_{k+1} =
A_{k+1} - (B_1 \union \ldots B_{k})$.  If $D$ is uncountable, we
must first well-order $D$; then a similar inductive construction
works.)
Thus, we have $A = A' \union (\union_{d\in D} B_d)$, and all the sets on
the right-hand side are pairwise disjoint.

From
(\ref{eqpl1}), it follows that $\Pl(A-A_d) > \Pl(A_d)$, for all $d \in
D$, so clearly $\Pl(A - B_d) > \Pl(B_d)$, since $B_d \subseteq A_d$.
{From} A2$^*$, it follows that
$\Pl(A') > \Pl(A - A')$.  Thus, $(\PL,w,v) \sat \phi \Cond \forall x
\psi$, as desired.  \eprf

\bigskip

\commentout{
\opro{crooked} The formula $\lottery \land \crooked$ is not
satisfiable in $\Ppossfo$ and $\Ppreffo$.
\eopro

\medskip

\prf Suppose $\POSS = (D,W,\Poss,\pi)$ is a possibility structure
such that
$\POSS \sat \lottery \land \crooked$.
Since $$\POSS \sat
\exists y\forall x
    (x \neq y \rimp
   ((\Winner(x) \lor \Winner(y)) \Cond \Winner(y))),$$
then there must
be some domain element $d \in D$ and valuation $v$ such that $v(y) =
d$ and
$$(\POSS,v) \sat \forall x (x \neq y \rimp \\
   ((\Winner(x) \lor \Winner(y)) \Cond \Winner(y))).$$
This in turn means that if
$v' \sim_x v$, then
\begin{equation}\label{eqcrooked}
(\POSS,w,v') \sat (\Winner(x) \lor \Winner(y)) \Cond \Winner(y).
\end{equation}
For each $d' \in D$, $v_{d'}$ be such that $v_{d'} \sim_x v$ and
$v_{d'}(x) = d'$.  Let $A_{d'} = \{w' \in W: (\Poss,w,v_{d'} \sat
\Winner(x)\}$.  It is immediate from (\ref{eqcrooked}) that
$\Poss(A_d) > \Poss(A_{d'} \inter \overline{A_d})$ if $d' \ne d$.  Thus,
it follows that
$$\Poss(A_d) \ge \Poss((\union_{d' \ne d} A_{d'}) \inter
\overline{A_d}).$$
(Note for future reference that here we are using the property that
if $\Poss(U) > \Poss(V_i)$, then $\Poss(U) \ge \Poss(\union_i V_i)$.)
Since
$\Poss(A_d) \ge \Poss((\union_{d' \ne d} A_{d'}) \inter A_d)$, it
follows immediately from the properties of possibility measures that
\begin{equation}\label{eqcrooked2}
\Poss(A_d) \ge \Poss(\union_{d' \ne d} A_{d'}).
\end{equation}

Since $(\POSS,w) \sat \forall x(\true \Cond \neg \Winner(x))$, we must
also have
\begin{equation}\label{eqcrooked3}
\Poss(\overline{A_d}) > \Poss(A_d).
\end{equation}
Putting together (\ref{eqcrooked2}) and (\ref{eqcrooked3}), we get that
$$\Poss(\overline{A_d}) > \Poss(\union_{d' \in D} A_{d'}).$$
Since, in general, if $\Poss(U) > \Poss(V)$, then $\Poss(U-V) >
\Poss(V)$, it follows that
$$\Poss(\overline{\union_{d' \in D} A_{d'}}) > \Poss(\union_{d' \in D}
A_{d'}).$$
However, that since $\POSS \sat \true \Cond \exists x \Winner(x)$,
we have that
$$\Poss(\union_{d' \in D} A_{d'}) > \Poss(\overline{\union_{d' \in
D} A_d}).$$
This gives us the desired contradiction.

The argument in the case of $\Ppreffo$ is similar in spirit.
Suppose $\PREF = (D,W,\prec,\pi)$ is a preferential structure that
$\PREF \sat \lottery \land \crooked$.  If $U, V \subseteq W$, following
\cite{Hal17}, we write
$V \prec U$ if for all $u \in U$, there exists some $v \in V$ such that
$v \prec u$ and there is no $u' \in U$ such that $u' \prec v$; we write
$V \preceq U$ if for all $u \in U$, there exists some $v \in V$ such
that $v \preceq u$.
It follows the definition of
$\Cond$ in preferential structures that
$(D,W,\prec,\pi) \sat \phi \Cond \psi$ iff $\intension{\phi \land \psi}
\prec \intension{\phi \land \neg \psi}$.
Moreover, it is easy to check that
\begin{enumerate}
\item If $U \prec V_i$, then $U \preceq \union_{i} V_i$
\item If $U \prec V$, then $U-V \prec V$.
\item If $U \prec V \preceq V'$, then $U \prec V'$.
\end{enumerate}
We can now essentially repeat
the proof above replacing $\Poss(U) > \Poss(V)$ by $U \prec V$ and
$\Poss(U)\ge \Poss(V)$ by $U \preceq V$, since the three properties
above were the only ones used in the proof.
} %

\opro{A2dagger}
 A2$^\dagger$ holds in every plausibility structure in $\Ppreffo$ and
 $\Ppossfo$.
\eopro
\prf
We start with $\Ppossfo$. Suppose that $\POSS = (D,W,\Poss,\pi)$ is a
possibility structure. Let
$\{A_i: i
\in I\}$ be a collection of pairwise disjoint sets such that
$\Poss(A_0) > \Poss(A_i)$ for all $i \in I - \{0\}$. This implies that
$\Poss(A_0) \ge \sup_{i \in I - \{0\}} \Poss(A_i) = \Poss(A - A_0)$. We
immediately get that $\Poss(A_0) \not< \Poss(A - A_0)$.

The argument in the case of $\Ppreffo$ is similar in spirit, although
somewhat more involved.
Suppose $\PL = (D,W,\Pl_{\prec},\pi) \in \Ppreffo$, where $\Pl_{\prec}$
is constructed from a partial order $\prec$ on $W$ as described
in Section~\ref{prop-logic}.  Again, let
$\{A_i: i
\in I\}$ be a collection of pairwise disjoint sets such that
$\Pl_\prec(A_0) > \Pl_\prec(A_i)$ for all $i \in I - \{0\}$.

By way of contradiction, suppose that $\Pl_{\prec}(A_0) <
\Pl_{\prec}(A - A_0)$. Let $w \in A_0$. Since
$\Pl_{\prec}(A_0) < \Pl_{\prec}(A - A_0)$, there is a world $w' \in
A-A_0$ such that $w' \prec w$ and there is no $w'' \in A_0$ such that
$w'' \prec w'$.  Let  $A_i$ be the set that contains
$w'$. (There must be such an index, since $A-A_0$ is the union of such
sets.) Since $\Pl_{\prec}(A_i) <
\Pl_{\prec}(A_0)$, there is a world $w'' \in A_0$ such that $w'' \prec
w'$. Thus, we get a contradiction. We conclude that $\Pl_{\prec}(A_0)
\not< \Pl_{\prec}(A - A_0)$ .
\eprf

\opro{A3*axioms}
The axiom $$\forall x \PBox \phi \rimp \PBox(\forall x \phi)$$ is sound
in  structures satisfying A3$^*$.  Moreover, the axiom
$$\forall x (\phi \Cond \psi) \rimp ((\exists x \phi) \rimp \psi),\ \ \mbox{if
$x$ does not appear free in $\psi$}
$$ is sound in structures satisfying A2$^*$ and A3$^*$.
\eopro
\prf
For the first part of the proposition, suppose that $\PL = (D,
W,\Pl,\pi)$ be a plausibility structure satisfying A3$^*$. Assume
that there is a world $w \in
W$ and a valuation $v$ such that
$$(\PL,w,v) \sat \forall x \PBox \phi.$$
This means that if $v' \sim_x v$, then
\begin{equation}\label{a3*-1}
(\PL,w,v') \sat \PBox\phi
\end{equation}
For each $d \in D$, let $v_{d}$ be
the valuation
such that $v_{d} \sim_x v$ and
$v_{d}(x) = d$. Let $A_d = \{ w' : (\PL,w',v_{d}) \sat \phi \}$. From
(\ref{a3*-1}), we have that $\Pl(W - A_{d}) = \bot$ for all $d \in
D$. Using A3$^*$, we get that $\PL(W - (\inter_d A_{d})) =
\bot$. Thus,
$$
(\PL,w,v) \sat \PBox \forall x \phi
$$
as desired.

For the second part of the proposition, suppose that $\PL = (D,
W,\Pl,\pi)$ be a plausibility structure satisfying A2$^*$ and A3$^*$.
Assume
that there is a world $w \in
W$ and a valuation $v$ such that
$$(\PL,w,v) \sat \forall x  (\phi \Cond \psi),$$ where $x$ does not
appear free in $\phi$.
This means that if $v' \sim_x v$, then
\begin{equation}\label{a3*-2}
(\PL,w,v') \sat \phi \Cond \psi.
\end{equation}
Again, for each $d \in D$,
let $v_{d}$ be
the valuation
such that $v_{d} \sim_x v$ and
$v_{d}(x) = d$. Let $A_d = \{ w' : (\PL,w',v_{d}) \sat \phi \}$ and let
$B = \{ w' : (\PL, w',v) \sat \psi \}$.

If $\Pl(A_d) = \bot$ for all $d$, then by A3$^*$ we get that
$\Pl(\union_d A_d) = \bot$. In this case $(\PL,w,v) \sat (\forall x
\phi) \Cond \psi$ vacuously.

On the other hand, if $\Pl(A_d) \ne \bot$ for all $d \in D$, then we
must have $\Pl(\union_{d \in D} A_d) > \bot$.
By (\ref{a3*-2}), for each $d' \in D$, we have that
either $\Pl(A_d) = \bot$ or $\Pl(A_d \inter B) > \Pl(A_d - B)$. (Note
that since $x$ is not free in $\psi$, we have that $(\PL,w',v) \sat
\psi$ iff
$(\PL,w',v_{d}) \sat \psi$.) In either case, we can conclude that $$
\Pl((\union_d A_d) \inter B) > \Pl(A_{d'} - B).
$$
Let $A'_d$ be pairwise disjoint sets such that $A'_d \subseteq A_d$,
and $\union_d A'_d = A_d$. (Such sets must exist; see the proof of
Proposition~\ref{forall3}.)
Thus, we have that $\Pl((\union_d A_d) \inter B) > \Pl(A'_d - B)$. Using
A2$^*$, we get that
$$
\Pl((\union_d A_d) \inter B) > \Pl(\union_d A'_d - B)= \Pl(\union_d
A_d - B).
$$
We conclude that
$$
(\PL,w,v) \sat (\forall x \phi) \Cond \psi,
$$
as desired.
\eprf

\opro{crooked} The formula $\lottery \land \crooked \rimp (\true
\Cond \false)$ true in structures satisfying A2$^\dagger$ and A3$^*$.
\eopro

\prf
Let $\PL = (D, W,\Pl,\pi)$ be a plausibility structure
satisfying A2$^\dagger$ and A3$^*$.
Suppose also that $\PL \sat \lottery \land
\crooked$.
Since $$\PL \sat
\exists y\forall x (x \neq y \rimp
   ((\Winner(x) \lor \Winner(y)) \Cond \Winner(y))),$$
there must
be some domain element $d_0 \in D$ and valuation $v$ such that $v(y) =
d_0$ and
$$
(\PL,v) \sat \forall x (x \neq y \rimp \\ ((\Winner(x) \lor
\Winner(y)) \Cond \Winner(y))).
$$
This in turn means that if
$v' \sim_x v$, then
\begin{equation}\label{eqcrooked}
(\PL,w,v') \sat (\Winner(x) \lor \Winner(y)) \Cond \Winner(y).
\end{equation}
For each $d \in D$, $v_{d}$ be such that $v_{d} \sim_x v$ and
$v_{d}(x) = d$.
Let $A_{d} = \{w' \in W: (\PL,w,v_{d} \sat \Winner(x)\}$, let
$A = \union_d A_{d}$, and let $B = W - A$.

It is immediate from (\ref{eqcrooked}) that either
\begin{itemize}\denselist
 \item [(a)] $\Pl(A_{d_0}) = \Pl(A_{d_0} - A_{d}) = \bot$ for
 all $d \neq d_0$, or
 \item[(b)] $\PL(A_{d_0}) > \PL(A_{d} - A_{d_0})$
for all $d \neg d_0$.
\end{itemize}
Assume that (a) is true. By A3$^*$, we get that $\Pl(\union_d A_d)
= \bot$. Since $\PL \sat \true \Cond \exists x \Winner(x)$, we get
that either $\Pl(W) = \bot$ or $\Pl(\union_d A_d) > \Pl(W -
(\union_d A_d))$. Since the latter
inequality is inconsistent, we conclude
that $\Pl(W) = \bot$ and, thus, $\PL \sat \true \Cond \false$, as
desired.

Now assume that (b) is true. From A2$^\dagger$,
it follows that
\beqn
\Pl(A_{d_0}) \not< \Pl(A - A_{d_0})
\label{crooked-1}
\eeqn

Since $\PL \sat \forall x (\true \Cond \neg\Winner(x))$, we have that
$$
(\PL, v_{d_0}) \sat \true \Cond \neg\Winner(x)
$$
Thus,
\beqn
\Pl(A_{d_0}) < \Pl(W - A_{d_0}) = \Pl(B \union (A - A_{d_0})).
\label{crooked-2}
\eeqn
Finally, since $\PL \sat \true \Cond \exists x \Winner(x)$, we have
that
\beqn
\Pl(B) < \Pl(A) = \Pl(A_{d_0} \union (A - A_{d_0})).
\label{crooked-3}
\eeqn
Using A2, (\ref{crooked-1}), and (\ref{crooked-2}), we get that
$\Pl(A - A_{d_0}) > \Pl(A_{d_0} \union B ) \ge \Pl(A_{d_0})$. This, however,
contradicts (\ref{crooked-1}), showing that (b) is impossible.
\eprf

\bigskip

\othm{statcompleteness}
$\SysCsfo$ is a sound and complete axiomatization of $\LCondsfo$
with respect to
$\PQPLsfo$. \eothm

\medskip
\prf We proceed much as in the proof of Theorem~\ref{subjcompleteness},
using the same notation.  Again we can assume without loss of
generality that $\phi$ is a sentence.  Using standard arguments of
first-order
logic, we can show that there is a $\C$-good maximal consistent set of
sentences $C^*$ that
includes $\phi$.  (The second property of $\C$-goodness, that $\forall
x \psi \in C^*$ implies $\psi[x/\boldc] \in C^*$, now follows
immediately
from the axioms, since we can substitute constants in arbitrary
contexts, and the proof of the first property is just the
standard first-order proof, again because we get to use all the standard
first-order axioms with no change.)

We construct a first-order statistical plausibility structure
$\PL = (\Dom, \Pl,\pi)$ by again taking
$\Dom
= \{[\boldc]: \boldc \in \C\}$ and defining $\pi$ so that
$\pi(\boldc) = [\boldc]$ for the constants $\boldc \in \C$ and $\pi$
for the symbols in $\Phi'$ is determined by the atomic sentences in
$C^*$ in the obvious
way.  The definition of $\Pl$ is also similar in spirit to that in the
proof of Theorem~\ref{subjcompleteness}.  We take the domain of $\Pl$ to
be all the definable subsets of $\Dom^\infty$.  More precisely, given a
formula $\phi$, let $\phi_v$ be the formula that results by replacing
all free occurrences of $x_i$ in $\phi$ by some constant in the
equivalence class $v(x_i)$.
(It doesn't matter which one.)  Let $A_\phi = \{v:
\phi_v \in C^*\}$.  The definable subsets are precisely those of the
form $A_\phi$ for some formula $\phi$.  Note that the definable subsets
do form an algebra, since $A_\phi \union A_\psi = A_{\phi \lor \psi}$
and $\overline{A_\phi} = A_{\neg \phi}$.  We define $\Pl$ on this
algebra by taking $\Pl(A_\phi) \le \Pl(A_\psi)$ iff $\SCond{\setx}{\phi
\lor \psi}{\psi} \in C^*$, where $\setx$ consists of all the variables
free in $\phi \lor \psi$.  We leave it to the reader to check
that for every sentence $\psi \in  \LCondsfo$, we have
$\PL \sat \psi$ iff $\psi \in C^*$.  \eprf

\subsection*{Acknowledgments}
We would like to thank Ronen Brafman, Ron Fagin, and Adam Grove
for their comments on an earlier version of the paper.

\bibliographystyle{chicago}
\bibliography{z,refs,bghk,conf-long}
\end{document}